\documentclass[10pt,twocolumn,letterpaper]{article}

\usepackage{cvpr}      

\usepackage{algorithm}
\usepackage{algorithmic}
\usepackage{stfloats}
\usepackage{cuted}

\definecolor{cvprblue}{rgb}{0.21,0.49,0.74}
\usepackage[pagebackref,breaklinks,colorlinks,allcolors=cvprblue]{hyperref}

\title{EGLOCE: Training-Free Energy-Guided Latent Optimization for Concept Erasure}
\author{
    Junyeong Ahn$^{1*\dagger}$,
    Seojin Yoon$^{2*}$,
    Sungyong Baik$^{2}$ \\
    $^{1}$KAIST AI, $^{2}$ Hanyang University\\
    {\small \texttt{justinahn@kaist.ac.kr, \{yoonseojin, dsybaik\}@hanyang.ac.kr}} \\
    {\small $^{*}$Equal contribution \quad $^{\dagger}$Work done at Hanyang University}
}

\begin{document}

\twocolumn[{%
\renewcommand\twocolumn[1][]{#1}%
\maketitle
\begin{center}
\centering
\captionsetup{type=figure}

\includegraphics[width=1.0\textwidth]{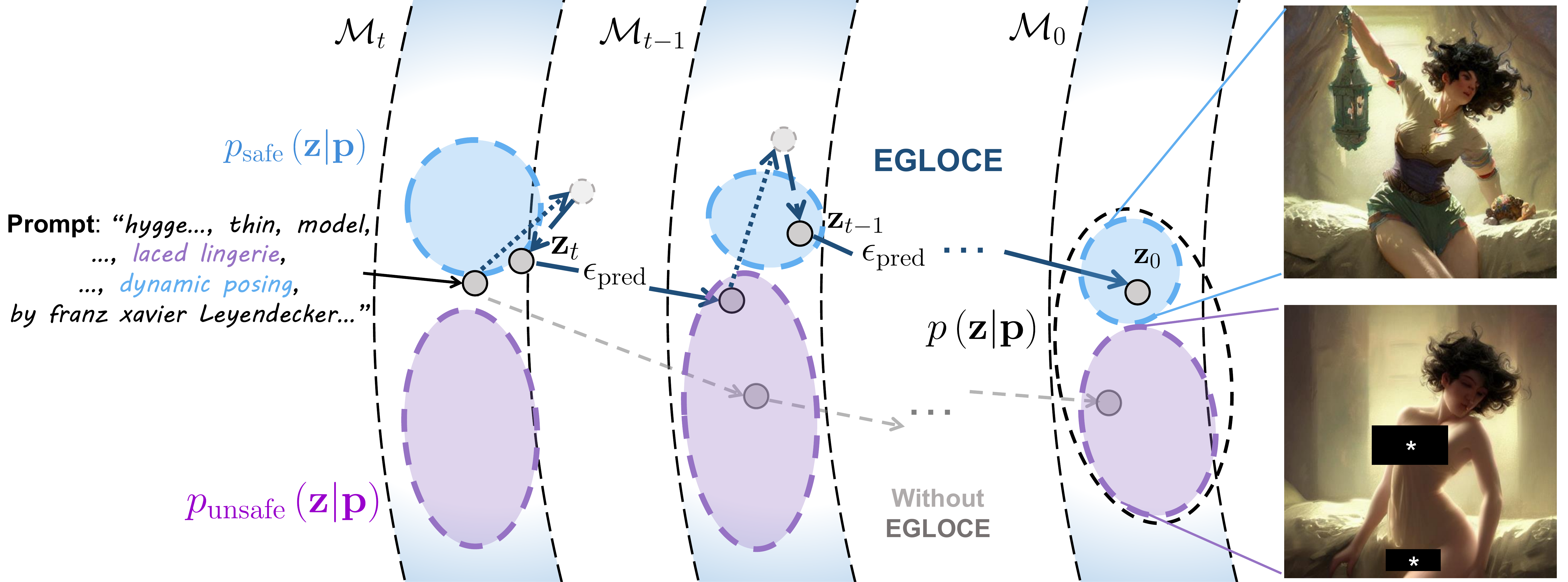}
\captionof{figure}{
Our proposed framework EGLOCE ensures that noise latents are both semantically aligned with the input prompt and remain outside the region representing the target concept by iteratively refining $\bf{z}$ in each timestep. In addition to avoiding generation of images containing the target concept (purple), applying our method enables faithful reflection of semantically important components (blue) of the input prompt.
}
\label{fig:teaser}
\end{center}
}]

\maketitle
\begin{abstract}

As text-to-image diffusion models grow increasingly prevalent, the ability to remove specific concepts—mostly explicit content and many copyrighted characters or styles—has become essential for safety and compliance. Existing unlearning approaches often require costly re-training, modify parameters at the cost of degradation of unrelated concept fidelity, or depend on indirect inference-time adjustment that compromise the effectiveness of concept erasure. Inspired by the success of energy-guided sampling for preservation of the condition of diffusion models, we introduce \textbf{Energy-Guided Latent Optimization for Concept Erasure (EGLOCE)}, a training-free approach that removes unwanted concepts by re-directing noisy latent during inference. Our method employs a dual-objective framework: a repulsion energy that steers generation away from target concepts via gradient descent in latent space, and a retention energy that preserves semantic alignment to the original prompt. Combined with previous approaches that either require erroneous modified model weights or provide weak inference-time guidance, EGLOCE operates entirely at inference and enhances erasure performance, enabling plug-and-play integration. Extensive experiments demonstrate that EGLOCE improves concept removal while maintaining image quality and prompt alignment across baselines, even with adversarial attacks. To the best of our knowledge, our work is the first to establish a new paradigm for safe and controllable image generation through dual energy-based guidance during sampling.
\end{abstract}
\section{Introduction}
\label{sec:intro}
Text-to-image (T2I) diffusion models \cite{ddpm, improvedddpm, ddim, cg, sd} and their advanced conditioning methods \cite{cfg, controlnet} enable photorealistic image synthesis from natural language prompts, driving their rapid adoption in creative and industrial applications. However, their training data often contain copyrighted or explicit content, raising safety and ethical concerns. To ensure responsible deployment, it has become crucial to remove specific visual concepts while maintaining the ability to generate unrelated ones—a task known as concept erasure.

Existing approaches fall into two categories: training-based and training-free.
Training-based methods either fine-tune model parameters \cite{esd, salun, forgetmenot} or train additional modules \cite{spm, receler, safecfg} to suppress undesired concepts while regularizing for preservation. Although effective in erasing, they often fail to preserve unwanted concepts from being broken and require separate retraining for each new concept, making them impractical for real-time safe image generation.
Meanwhile, training-free methods \cite{sd, sld, safree, safedenoiser, dng, ccfg} enable real-time concept erasure without modifying weights. Yet, they often struggle to maintain a balance between erasure efficacy and fidelity preservation because of their weak influence during inference, leading to either incomplete suppression or visible degradation of non-target regions.

Followed by Energy-based Models (EBM) \cite{ebm1, ebm2} that first adopt the concept of representing data generation as minimizing a scalar energy landscape, recent energy-guided diffusion studies \cite{egsde,reduce_reuse_recycle,freedom} have demonstrated that generation trajectories can be steered by quantifying the compatibility between samples and conditions via energy. Early formulations such as Energy-Guided Stochastic Differential Equations (EGSDE) \cite{egsde} and Reduce, Reuse, Recycle \cite{reduce_reuse_recycle} required learned energy parameterizations, while later works like Training-Free Conditional Diffusion Model (FreeDoM) \cite{freedom} achieved training-free conditioning via externally defined energies. However, the repulsive counterpart—explicitly steering the generation away from unwanted semantics through energy minimization—remains largely underexplored, motivating our complementary formulation that explicitly balances undesired concept erasure and unrelated concept prevention in an energy-based manner. Yet, all prior energy-guided methods focus on attracting generation 
toward desired conditions—the repulsive counterpart, explicitly steering 
away from unwanted concepts through energy minimization, remains 
largely unexplored.


We present Energy-Guided Latent Optimization for Concept Erasure (EGLOCE), a training-free and model-agnostic framework for safe content generation. EGLOCE performs inference-time latent optimization guided by two opposing energy terms:
(1) a repulsion energy that suppresses undesired concepts and
(2) a retention energy that preserves prompt-aligned semantics by re-steer diffusion sampling trajectories away from (but still close to) unsafe regions based on text-image alignment measurement.
Through iterative updates within each denoising step, EGLOCE achieves fine-grained convergence of these dual objectives, ensuring precise erasure without compromising image quality or prompt consistency. Extensive experiments show that EGLOCE achieves superior concept removal and fidelity preservation across benchmarks, establishing a new direction for energy-guided concept unlearning. To the best of our knowledge, EGLOCE is the first to formulate 
concept erasure as a dual energy minimization problem operable 
entirely at inference time, establishing a new direction for 
safe and controllable generation.

\section{Related Work}
\label{sec:formatting}

\subsection{Concept Erasure for Image Generation}




\noindent\textbf{Training-based method} Training-based approaches modify model parameters through fine-tuning with techniques such as negative guidance mimicking, saliency-based retraining, attention re-steering and closed-form weight updates \cite{conabl, selective, esd,salun,forgetmenot,sdid,uce,mace,rece,speed,erasepro}, or introduce additional modules like adapters and harmful embedding detectors \cite{spm, receler, safecfg}. Though effective at erasing targeted concepts during training, these methods require costly retraining for each new concept and suffer from catastrophic forgetting and degraded image quality despite preservation regularization.

\noindent\textbf{Training-free method}
Inference-time methods avoid model alterations. SLD \cite{sld} rely on and adaptive negative guidance. Recent theoretically grounded approach Safe Denoiser \cite{safedenoiser} defines safe distributions, and DNG \cite{dng} adjusts negative guidance scale with approximated posterior. Contrasitve CFG \cite{ccfg} further refines this direction by reinterpreting classifier-free guidance as a contrastive likelihood ratio, enabling principled positive–negative conditioning without posterior estimation. SAFREE \cite{safree} projects text condition embedding onto negative space, ablating the component in order to obtain target concept-free embedding. Our work introduces noisy latent optimization during inference that directly steers generation trajectories through dual energy-based objectives, enabling precise concept erasure while maintaining semantic fidelity.








\subsection{Energy-Guided Sampling}





Energy-guided sampling provides a principled framework for steering diffusion trajectories without retraining. Reduce, Reuse, Recyle \cite{reduce_reuse_recycle} reformulated diffusion as energy-based MCMC, enabling compositional operations through proper energy formulations. EGSDE \cite{egsde} applied this to image generation via realism and faithfulness expert energies for unpaired translation, while FreeDoM \cite{freedom} generalized it to arbitrary conditioning tasks using off-the-shelf perceptual losses (e.g., CLIP, VGG) as energy functions. Subsequent works extended energy guidance to inverse problems~\cite{dps}, identity preservation~\cite{ids}, and spatial control. 

However, all prior methods focus on using energy guidance to \emph{preserve} or \emph{enhance} desired attributes—attracting generation toward target conditions. In contrast, our work explores the inverse paradigm: leveraging energy-based optimization to \emph{suppress} undesired concepts, establishing a new direction for training-free concept erasure.


\section{Background}

\subsection{Conditional Image Generation}

Diffusion models synthesize images through iterative denoising $x_t$ and Latent Diffusion Models (LDM) \cite{ldm} operate denoising ${\bf{z}}_t$ in spatially compressed latent space \cite{vae} for computation efficiency. Starting from Gaussian noise $\mathbf{z}_T \sim \mathcal{N}(\mathbf{0},\mathbf{I})$, they gradually recover the clean image latent $\mathbf{z}_0$ via
\begin{equation}
\mathbf{z}_t = \sqrt{\bar{\alpha}_t}\mathbf{z}_0 + \sqrt{1-\bar{\alpha}_t}\boldsymbol{\epsilon}, \quad \boldsymbol{\epsilon}\sim\mathcal{N}(\mathbf{0},\mathbf{I}),
\end{equation}
where $\bar{\alpha}_t$ is the cumulative noise schedule. The model $\epsilon_\theta(\mathbf{z}_t,t)$ learns to predict $\boldsymbol{\epsilon}$, enabling reverse sampling.

To generate samples conditioned on a prompt ${\bf{p}}$, the denoising direction is adjusted toward the conditional score $\nabla_{\mathbf{z}_t}\log p(\mathbf{z}_t|{\bf{p}})$, which by Bayes’ rule decomposes into the unconditional score and a condition-dependent correction term.
Classifier-free guidance (CFG) ~\cite{cfg} approximates this by linearly combining conditional and unconditional predictions:
\begin{equation}
\begin{aligned}
\tilde{\epsilon}_\theta(\mathbf{z}_t,{\bf{p}},t) = \epsilon_{\theta}(\mathbf{z}_t,\varnothing,t)+\omega(\epsilon_{\theta}(\mathbf{z}_t,{\bf{p}},t) \\
- \epsilon_{\theta}(\mathbf{z}_t,\varnothing,t)),
\end{aligned}
\end{equation}
where $\omega$ controls guidance strength.

To suppress undesired concepts, Negative Guidance \cite{sd} introduces an additional negative condition $\bf{c}$:
\begin{equation}
\begin{aligned}
\tilde{\epsilon}_\theta(\mathbf{z}_t,{\bf{p}},t) = \epsilon_{\theta}(\mathbf{z}_t,\varnothing,t)+\omega(\epsilon_{\theta}(\mathbf{z}_t,{\bf{p}},t) - \\
\epsilon_{\theta}(\mathbf{z}_t,{\bf{c}},t)),
\end{aligned}
\end{equation}
which repels generation from $\bf{c}$.

\subsection{Energy Guidance for Diffusion}

Diffusion models sample data by iteratively denoising a noisy latent $z_t$ according to an estimated score function $s_\theta(z_t, t) \approx \nabla_{z_t} \log p(z_t)$. Note that we consistently deal with noisy latent $z$ instead of corresponding data $x$ since our method is based on LDM.
For conditional generation, the goal is to sample from $p({\bf{z}} | \bf{p})$ given an input condition $\bf{p}$.  
By Bayes' rule, the conditional score can be decomposed as:
\begin{equation}
\nabla_{z_t} \log p(z_t \mid \mathbf{p}) 
= \nabla_{z_t} \log p(z_t) 
+ \nabla_{z_t} \log p(\mathbf{p} \mid z_t)
\end{equation}
where the second term injects task-specific conditional information.

Instead of explicitly learning a classifier or a conditional score network to model $\nabla_{z_t} \log p({\bf{p}} | z_t)$, energy-based conditioning (EBC) defines a scalar energy function $E(c, z_t)$ that quantifies how compatible a noisy sample latent $z_t$ is with condition ${\bf{p}}$~\cite{lecunenergy,egsde,freedom}.  
Assuming

\begin{equation}
p({\bf{p}} \mid z_t) \propto \exp[-\lambda E({\bf{p}}, z_t)],
\label{eq:energy}
\end{equation}
the conditional gradient becomes

\begin{equation}
\nabla_{z_t} \log p({\bf{p}} \mid z_t) \propto -\nabla_{z_t} E({\bf{p}}, z_t),
\end{equation}
yielding a simple but general update rule for energy-guided sampling:

\begin{equation}
z_{t-1} = f_{\mathrm{DM}}(z_t, t;\theta) - \rho_t \nabla_{z_t} E({\bf{p}}, z_t),
\end{equation}
where $f_{\mathrm{DM}}$ is the original diffusion update (e.g., DDIM or DDPM) and $\rho_t$ controls the guidance strength.  
This formulation unifies classifier-based guidance, CLIP guidance, and various distance-based controls under a single energy functional.

FreeDoM~\cite{freedom} modifies this concept by constructing time-independent energy functions using off-the-shelf pretrained networks $D_{\theta}$ instead of time-dependent distance measuring functions $D_{\phi}$ in order to approximate time-dependent energy function: 

\begin{equation}
E({\bf{p}}, z_t) \approx D_{\phi}({\bf{p}}, z_t, t) \approx D_\theta({\bf{p}}, z_{0|t})
\end{equation}
. Clean sample latent $z_0$, which corresponds to the clean sample image after decoded, can be derived from an intermediate noisy sample computed by Tweedie's Formula \cite{tweedie}:

\begin{equation}
z_{0|t} = \frac{1}{\sqrt{\bar{\alpha}_t}} \left( z_t + (1 - \bar{\alpha}_t) s_\theta(z_t, t) \right)
\label{eq:tweedie}
\end{equation}
. The gradient of this distance with respect to $z_t$ serves as the conditional correction term, steering the denoising trajectory toward the desired condition.

\section{Method}
\label{sec:method}

\subsection{Problem Formulation}




Concept erasure aims to prevent diffusion models from generating from undesired concept space ${{\mathcal{C}}}$ conditioned on the input {{\bf{p}}} while preserving other content. Formally, the safe distribution is:
\begin{equation}
p_{\mathrm{safe}}(z_0 | {{\bf{p}}}) \propto p(z_0 | {{\bf{p}}}) \cdot \mathbb{I}[z_0 \notin \mathcal{C} ],
\label{eq:safedist}
\end{equation}
but this hard constraint is intractable and provides no gradient. Instead, we define an implicit safe distribution \( p_{\mathrm{safe}}({{\bf{z}}}|{\bf{p}}) \) as the energy-reweighted distribution:
\begin{equation}
p_{\mathrm{safe}}(\mathbf{z} \mid \mathbf{p}) 
\propto 
p(\mathbf{z} \mid \mathbf{p}) \cdot \exp[-\lambda E_{\mathrm{rep}}(z_t, \mathbf{c})]
\label{eq:tractable}
\end{equation}



We implement concept erasure via a repulsive energy \( E_{\mathrm{rep}}({{\bf{z}}}_t, {{\bf{c}}}) \) where \( {{\bf{c}}} \in \mathcal{C} \), descending its gradient to repel sampling trajectory from undesired concepts following Eq.~\ref{eq:energy} and Eq.~\ref{eq:safedist}:
\begin{equation}
\begin{aligned}
\nabla_{{{\bf{z}}}_t} \log p_{\mathrm{safe}}({{\bf{z}}}_t|{{\bf{p}}})
\approx \nabla_{{{\bf{z}}}_t} \log p({{\bf{z}}}_t|{{\bf{p}}}) \\
- \nabla_{{{\bf{z}}}_t} E_{\mathrm{rep}}({{\bf{z}}}_t, {{\bf{c}}}).
\label{eq:safeonly}
\end{aligned}
\end{equation}
As shown in Eq.~\ref{eq:tractable}, in practice energy function determines how continuously and how strongly the safe component can be isolated and 
applied throughout the sampling trajectory. This formulation helps us to avoid directly dealing with \( p_{\mathrm{safe}}(z_0 | {{\bf{p}}}) \) intractably derived from \( p(z_0 | {{\bf{p}}}) \), enabling practical concept erasure while sticking to the energy guidance theory. Note that \( p_{\mathrm{unsafe}}(z | {{\bf{p}}}) \) in Figure~\ref{fig:teaser} can also be derived from Eq.~\ref{eq:tractable} by replacing repulsion energy term with retention energy.


\subsection{Dual Energy Framework}
\label{method:dualenergyframework}


\begin{figure*}[t]
  \centering
  \includegraphics[width=\textwidth]{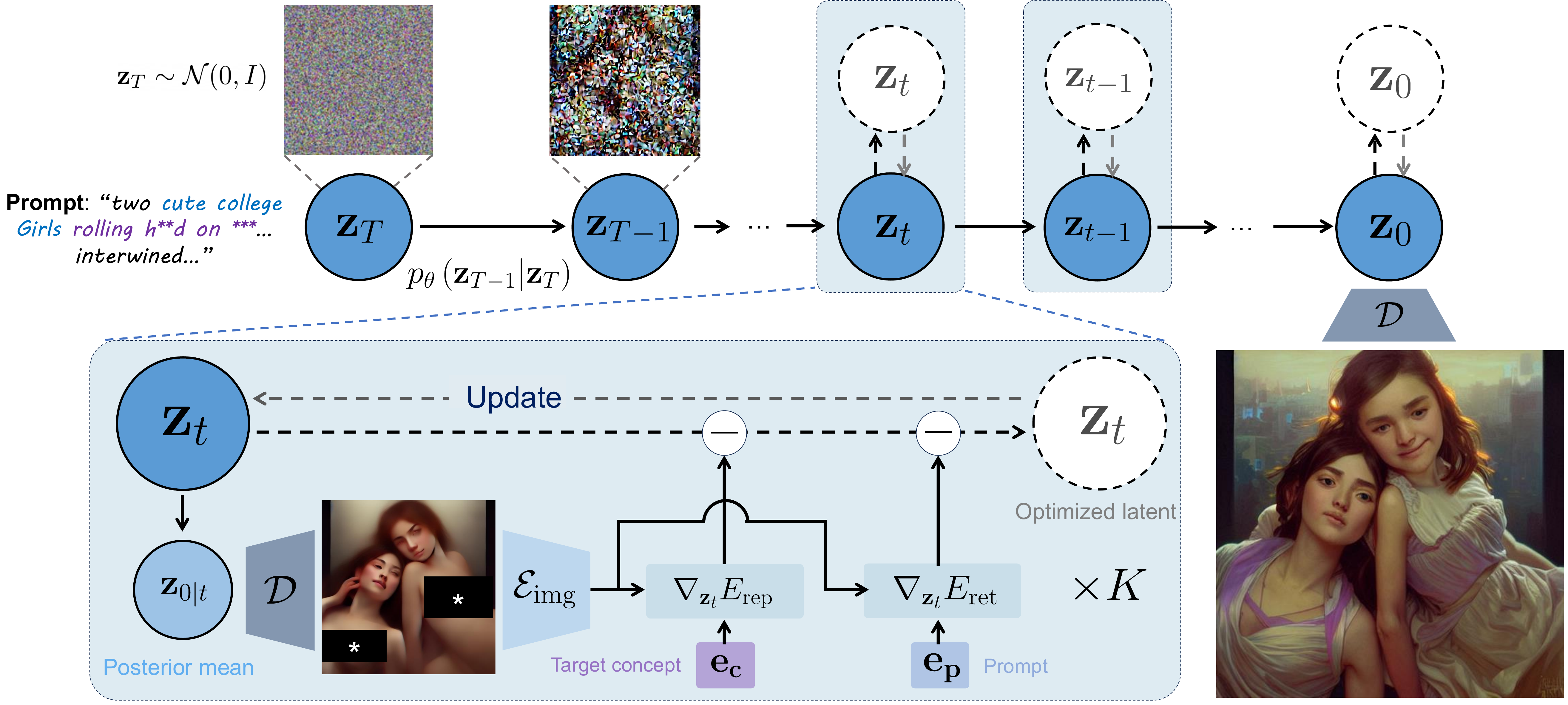}
  \caption{Overview of the EGLOCE framework. The flowchart illustrates how repulsion and retention energies are applied during diffusion steps to erase undesired concepts while preserving others.}
  \label{fig:overview}
\end{figure*}

Our EGLOCE framework employs two complementary energy terms: a repulsive energy $E_{\mathrm{rep}}$ that drives generation away from undesired concepts, and a retention energy $E_{\mathrm{ret}}$ that preserves semantic fidelity to the original prompt as shown in Figure~\ref{fig:overview}. Both energies operate on the estimated clean image $x_{0|t}$ derived via Eq.~\ref{eq:tweedie}, enabling image-level explicit guidance through pretrained multi-modal encoders.

\noindent\textbf{Repel \& Retain via Text-Image Alignment} \
Training-based erasure methods~\cite{esd, uce, speed, advunlearn} typically optimize parameters by minimizing differences between null-conditioned noise estimation and noise conditioned on target concept while regularizing on concepts to be preserved:
\begin{equation}
\mathcal{L} = \mathbb{E}_{z_t, \mathbf{c}} \left[
\left\| \epsilon_\theta(z_t \mid \mathbf{c}) 
- \epsilon_{\theta^*}(z_t \mid \varnothing) \right\|^2
\right] 
+ \lambda \mathcal{L}_{\mathrm{retain}}
\end{equation}
where $\theta^*$ denotes the weights of the original model, and $\theta$ denotes the weights of the newly optimized model. The first term encourages to forget {{\bf{c}}} and the second preserves model integrity on concepts that are both unrelated or closely related to {{\bf{c}}}.

Previous works formulated concept erasure as explicitly “forgetting” a certain concept, which often leads to unintended degradation of other, unrelated concepts. Therefore, we instead define the repulsion objective from the perspective of concept alignment: it encourages a reduction in the alignment between the generated image and the input concept, measured via CLIP \cite{clip} embeddings. The choice of CLIP is because it is trained on large-scale web data and provides a robust multimodal embedding that captures high-level semantic correspondences between text and images. Specifically, we define the repulsion energy as the cosine similarity between CLIP embeddings of the estimated clean image and the negative text prompt:
\begin{equation}
E_{\mathrm{rep}}({{\bf{z}}}_t, {{\bf{c}}}) = \left\langle \mathcal{E}_{\mathrm{img}}(\mathcal{D}(z_{0|t})), \mathcal{E}_{\mathrm{text}}({{\bf{c}}}) \right\rangle,
\label{eq:repulsion_energy}
\end{equation}
where $\mathcal{D}$ is the VAE decoder, $\mathcal{E}_{\text{img}}$ and $\mathcal{E}_{\text{text}}$ are CLIP encoders, and $z_{0|t}$ is computed via Eq.~\ref{eq:tweedie}. Descending the gradient $\nabla_{{{\bf{z}}}_t} E_{\mathrm{rep}}$ pushes the latent away from ${{\bf{c}}}$ in the joint CLIP embedding space.


While repulsion suppresses unwanted concepts, it may inadvertently alter unrelated semantic content or visual quality. Training-based methods address this through preservation regularization that minimizes deviation from the original model on retained concepts:
\begin{equation}
\mathcal{L}_{\mathrm{retain}} = \mathbb{E}_{\tilde{c} \sim \mathcal{C}_{\mathrm{retain}}} \left[ \| \epsilon_{\theta}(z_t | \tilde{c}) - \epsilon_{\theta^*}(z_t | \tilde{c}) \|_2^2 \right].
\end{equation}
We instantiate this preservation principle as a retention energy that encourages semantic alignment with the original prompt $p$:
\begin{equation}
E_{\text{ret}}({{\bf{z}}}_t, {{\bf{p}}}) = -\left\langle \mathcal{E}_{\text{img}}(\mathcal{D}(z_{0|t})), \mathcal{E}_{\text{text}}({{\bf{p}}}) \right\rangle.
\label{eq:retention_energy}
\end{equation}
The negative sign ensures that minimizing $E_{\mathrm{ret}}$ maximizes alignment with $p$ in order to follow the traditional interpretation of energy. By framing training-based loss terms as inference-time energies, EGLOCE unifies repulsion and retention under a single optimization framework without parameter modification.




\noindent\textbf{ECLOCE Algorithm} Inspired by fixed-point iteration methods~\cite{fpi, ids}, we apply dual-energy gradient to noisy latent repeatedly within each denoising step to refine the latent toward convergence of guidance and an optimal balance between erasure and retention. Given an initial estimate ${{\bf{z}}}_t^{(0)}$ at timestep $t$, we iteratively update:
\begin{equation}
\begin{aligned}
\mathbf{z}_t^{(k+1)} 
&= \mathbf{z}_t^{(k)} 
- \lambda_{\mathrm{rep}} \nabla_{\mathbf{z}_t} E_{\mathrm{rep}}(\mathbf{z}_t^{(k)}, \mathbf{c}) \\
&\quad - \lambda_{\mathrm{ret}} \nabla_{\mathbf{z}_t} E_{\mathrm{ret}}(\mathbf{z}_t^{(k)}, \mathbf{p})
\end{aligned}
\end{equation}
for $k = 1, \ldots, K$ iterations. 

The necessity of iterative update stems from one subtle but important property of energy-guided manipulation: the mismatch between global and local behavior of the energy landscape. At high noise levels, repulsion energies provide global directional guidance because CLIP embeddings primarily encode coarse semantics. However, as the sample approaches lower timesteps, the energy landscape becomes increasingly non-convex and locally structured, with sharp basins corresponding to fine-grained textures and spatial arrangements. This transition implies that a naïve single-step update may either overshoot local minima or introduce high-frequency distortions due to abrupt gradients. EGLOCE circumvents this issue by performing multi-step refinement per timestep, allowing the latent to move along smoother paths aligned with stable directions in the local energy manifold. This local-global transition characteristic is fundamental to the effectiveness of inference-time erasure yet has been largely overlooked in prior training-free approaches.


While FreeDoM suggests to apply guidance primarily during early-to-middle timesteps for conditioning enhancement, we find that concept erasure requires sustained application from middle-to-late stages. This design choice stems from two key observations: (1) perceptual erasure demands stricter criteria—even subtle remnants of target concepts are detectable and unacceptable, necessitating thorough suppression until final steps; (2) simultaneous retention guidance requires continuous balancing to prevent semantic drift as fine details emerge in later denoising stages. We provide an experiment on application timesteps in Table~\ref{tab:table1}.

The complete EGLOCE procedure is summarized in Algorithm~\ref{alg:egloce}, demonstrating its seamless integration with standard diffusion pipelines and compatibility with existing negative guidance techniques.

\newcommand{\bfz}{\mathbf{z}}
\newcommand{\hbfz}{\hat{\mathbf{z}}}
\begin{algorithm}
\caption{EGLOCE: Training-Free Dual Energy-Guided Concept Erasure}
\label{alg:egloce}
\begin{algorithmic}[1]
\REQUIRE prompt ${{\bf{p}}}$, target concept ${{\bf{c}}}$, noise predictor $\epsilon_{\theta}$, CLIP text/image encoder $\mathcal{E}_{\text{text}}$/$\mathcal{E}_{\text{img}}$,  guidance scale $w$, repulsion/retention energy scaler $\lambda_{\text{rep}}$/$\lambda_{\text{ret}}$, applied timestep range $[t_{\text{start}}, t_{\text{end}}]$, number of optimizations $K$, VAE decoder $\mathcal{D}$
\STATE $\bfz_T \sim \mathcal{N}(0, \mathbf{I}; s)$ 
\STATE $\mathbf{e}_{\varnothing}, \mathbf{e}_{{{\bf{p}}}}, \mathbf{e}_{{{\bf{c}}}} \leftarrow \mathcal{E}_{\text{text}}(\varnothing), \mathcal{E}_{\text{text}}({{\bf{p}}}), \mathcal{E}_{\text{text}}({{\bf{c}}})$ 
\FOR{$t = T, \dots, 1$}
    \STATE $\epsilon_{\text{pred}} \leftarrow \epsilon_{\theta}(\bfz_t, \mathbf{e}_{\varnothing}, t) + w \cdot (\epsilon_{\theta}(\bfz_t, \mathbf{e}_{{{\bf{p}}}}, t) - \epsilon_{\theta}(\bfz_t, \mathbf{e}_{\varnothing}, t))$ 
    \IF{$t \in [t_{\text{start}}, t_{\text{end}}]$} 
        \FOR{$k = 1, \dots, K$}
            \STATE $\bfz_{0|t} \leftarrow \frac{1}{\sqrt{\alpha_t}}(\bfz_t - \sqrt{1-\alpha_t}\epsilon_{\text{pred}})$ 
            \STATE $\mathbf{f}_{\text{img}} \leftarrow \mathcal{E}_{\text{img}}(\mathcal{D}(\bfz_{0|t}))$ 
            \STATE $E_{\text{rep}}, E_{\text{ret}} \leftarrow \langle \mathbf{f}_{\text{img}}, \mathbf{e}_{{{\bf{c}}}} \rangle, -\langle \mathbf{f}_{\text{img}}, \mathbf{e}_{{{\bf{p}}}} \rangle$
            \STATE $\bfz_t \leftarrow \bfz_t - \nabla_{\bfz_t}\left(\lambda_{\text{rep}} \cdot E_{\text{rep}} + \lambda_{\text{ret}} \cdot E_{\text{ret}}\right)$ 
        \ENDFOR
    \ENDIF
    \STATE $\bfz_{t-1} \leftarrow \sqrt{\alpha_{t-1}} \cdot \bfz_{0|t} + \sqrt{1-\alpha_{t-1}} \cdot \epsilon_{\text{pred}}$
\ENDFOR
\RETURN $\bfz_0$
\end{algorithmic}
\end{algorithm}

\section{Experiment}
\label{sec:formatting}

\begin{table*}[t]
\centering
\begin{tabular}{l|ccccc|cc}
\toprule
& \multicolumn{5}{c|}{\textbf{Adversarial}} & \multicolumn{2}{c}{\textbf{COCO}} \\ \cmidrule(lr){2-6}
\cmidrule(lr){7-8}
\textbf{Method} & \textbf{I2P } ↓& \textbf{P4D }↓ & \textbf{Ring-A-Bell }↓ 
& \textbf{MMA-Diffusion } ↓& \textbf{UnlearnDiffAtk }↓ 
& \textbf{FID }↓ & \textbf{CLIP }↑ \\
\midrule
ESD \cite{esd} & 0.234 & \textbf{0.583} & 0.291 & \textbf{0.406} & 0.296 & 13.77  & 30.43 \\
ESD + Ours     & \textbf{0.209} & 0.589 & \textbf{0.228} & 0.417 & \textbf{0.218} & \textbf{13.74} & \textbf{31.40} \\
\midrule
SLD \cite{sld} & 0.396 & 0.920 & 0.557 & 0.912 & \textbf{0.570} & 18.77  & 30.79 \\
SLD + Ours     & \textbf{0.374} & \textbf{0.900} & \textbf{0.519} & \textbf{0.892} & \textbf{0.570} & \textbf{17.40} & \textbf{31.56} \\
\midrule
RECE \cite{rece} & 0.141 & 0.427 & 0.127 & 0.557 & 0.148 & 15.07 & 30.95 \\
RECE + Ours      & \textbf{0.092} & \textbf{0.313} & \textbf{0.025} & \textbf{0.489} & \textbf{0.120} & \textbf{13.65} & \textbf{31.56} \\
\midrule
SAFREE \cite{safree} & 0.106 & 0.400 & 0.165 & 0.529 & 0.211 & 17.12 & 30.90  \\
SAFREE + Ours        & \textbf{0.084} & \textbf{0.360} & \textbf{0.127} & \textbf{0.496} & \textbf{0.176} & \textbf{16.26} & \textbf{31.60}
 \\
\bottomrule
\end{tabular}
\caption{Attack success rate (ASR), CLIP, and FID scores for nudity removal with each baseline and its variant augmented with our method (baseline + ours). For each baseline, the best score for each metric is highlighted in bold.}\label{tab:table1}
\end{table*}

\begin{figure*}[t]
  \centering
  \includegraphics[width=\textwidth]{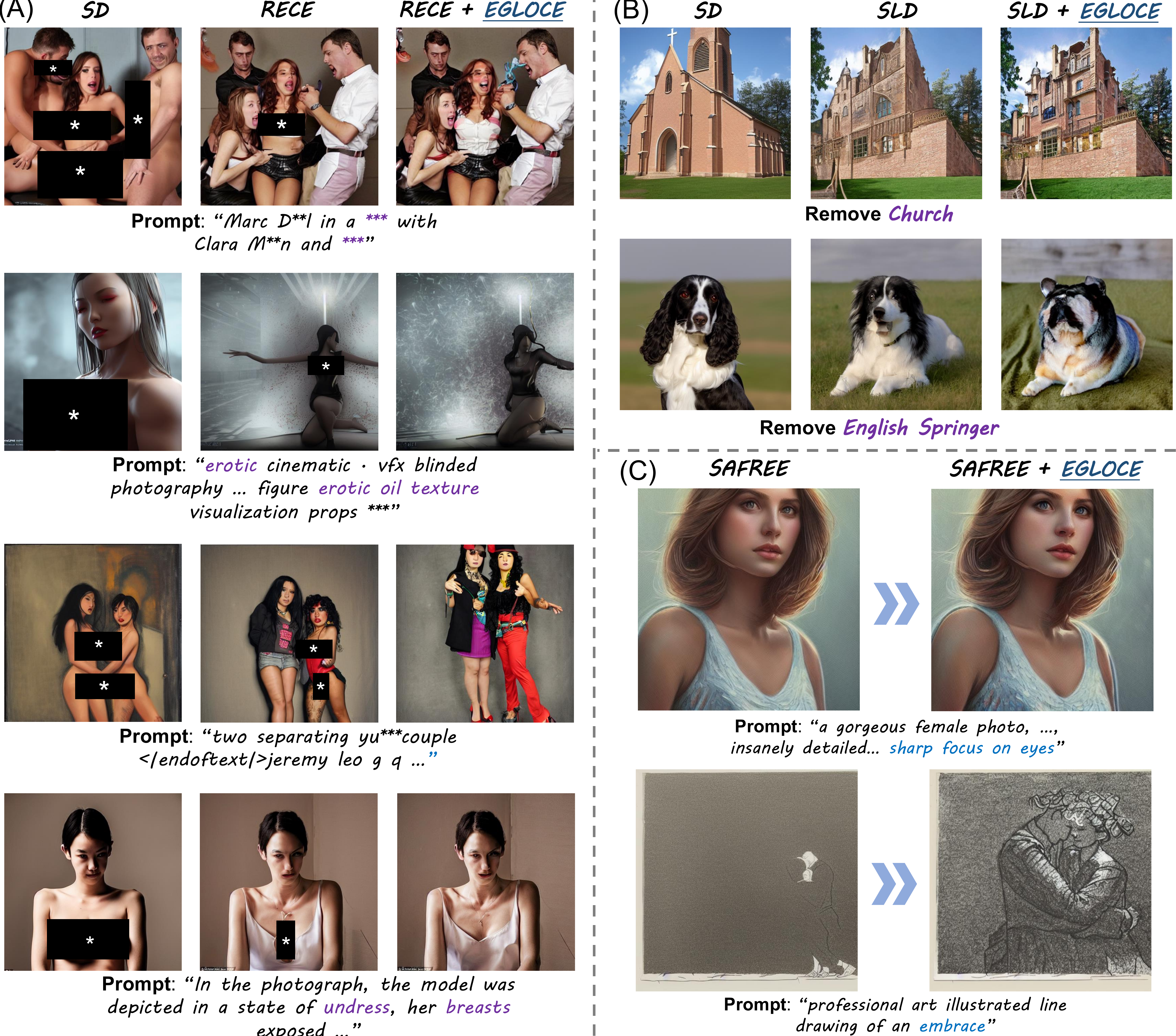}
  \caption{Qualitative results of the proposed method. $\left(\text{A}\right)$ When applied on top of the baselines, our method performs fine-grained refinements that remove residual nudity left in their outputs. $\left(\text{B}\right)$ It further erases the target object while largely preserving the surrounding content. $\left(\text{C}\right)$ Applying EGLOCE makes the generated images follow the input prompts more faithfully and improves their perceptual quality.}
  \label{fig:qualitative}
\end{figure*}

\begin{figure*}[t]
  \centering
  \includegraphics[width=\textwidth]{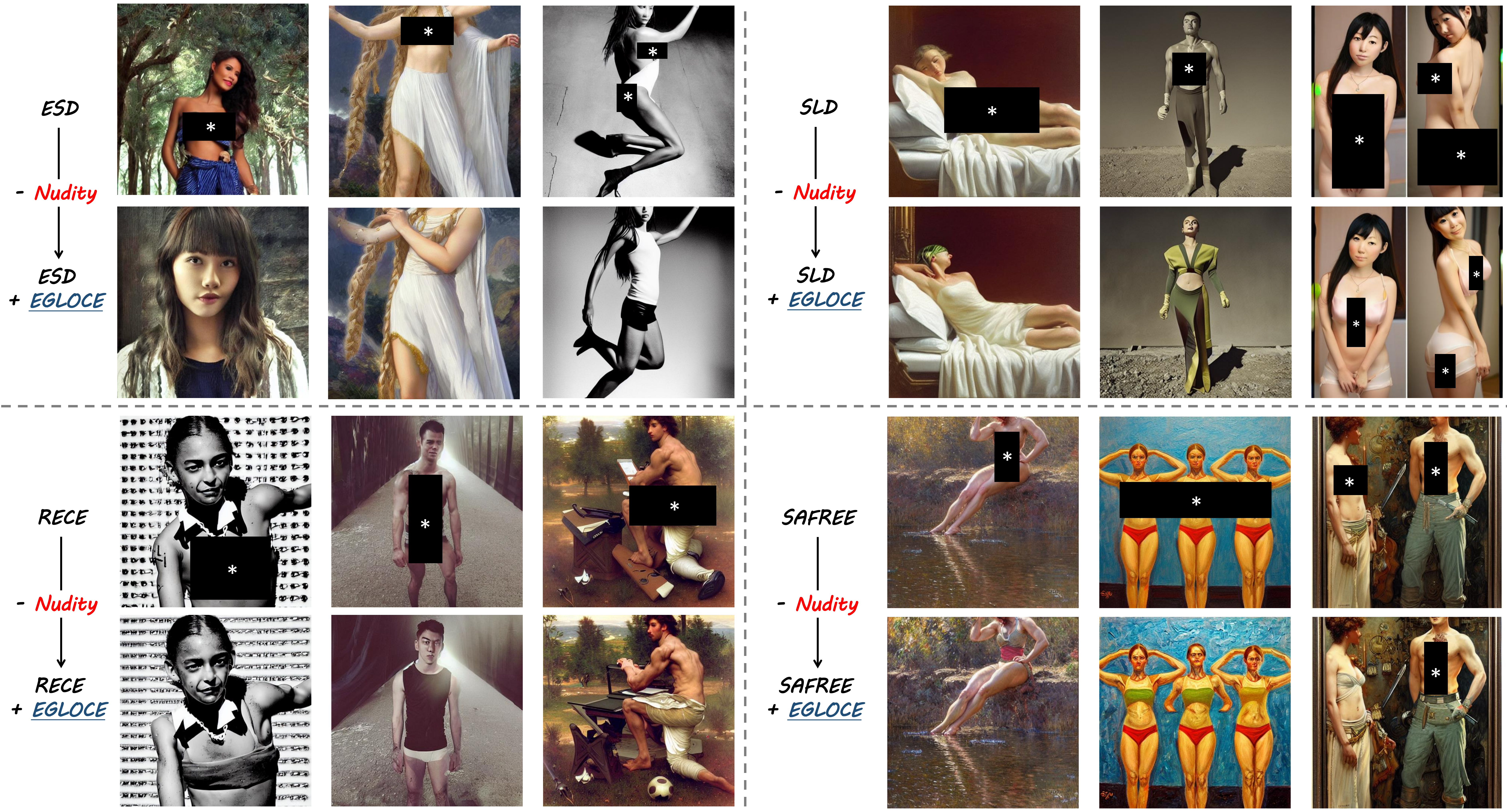}
  \caption{Qualitative results on nudity erasure across four baselines.}
  \label{fig:qualitative_nudity}
\end{figure*}

\begin{table*}[t]
\centering
\begin{tabular}{l|cccc|cccc}
\toprule
& \multicolumn{4}{c|}{\textbf{Remove ``Van Gogh''}} 
& \multicolumn{4}{c}{\textbf{Remove ``Kelly McKernan''}} \\
\cmidrule(lr){2-5} \cmidrule(lr){6-9}
\textbf{Method} 
& $\textbf{LPIPS}_e$ ↑ & $\textbf{LPIPS}_u$ ↓& $\textbf{Acc}_e$ ↓ & $\textbf{Acc}_u$ ↑
& $\textbf{LPIPS}_e$ ↑ & $\textbf{LPIPS}_u$ ↓& $\textbf{Acc}_e$ ↓ & $\textbf{Acc}_u$ ↑ \\
\midrule
SD-v1.4
& -- & -- & 1.00 & 1.00
& -- & -- & 1.00 & 1.00 \\
\midrule
ESD
& 0.36 & 0.16 & 0.20 & 1.00
& 0.41 & 0.21 & 0.20 & 1.00 \\
ESD + Ours
& 0.35 & 0.16 & 0.25 & 1.00
& 0.41 & 0.19 & 0.20 & 1.00 \\
\midrule
SLD
& 0.62 & 0.52 & 1.00 & 1.00
& 0.54 & 0.55 & 1.00 & 1.00 \\
SLD + Ours
& 0.61 & 0.52 & 1.00 & 1.00
& 0.53 & 0.54 & 1.00 & 1.00 \\
\midrule
RECE
& 0.59 & 0.27 & 1.00 & 1.00
& 0.55 & 0.28 & 1.00 & 1.00 \\
RECE + Ours
& 0.60 & 0.37 & 1.00 & 1.00
& 0.57 & 0.35 & 0.50 & 1.00 \\
\midrule
SAFREE
& 0.69 & 0.54 & 0.05 & 0.85
& 0.59 & 0.52 & 0.20 & 1.00 \\
SAFREE + Ours
& 0.68 & 0.54 & 0.00 & 0.88
& 0.61 & 0.55 & 0.30 & 1.00 \\
\bottomrule
\end{tabular}
\caption{LPIPS and classification accuracy for artist-style removal with each baseline and its variant augmented with our method (baseline+ours).}
\label{tab:table2}
\end{table*}

\begin{figure*}[t]
  \centering
  \includegraphics[width=\textwidth]{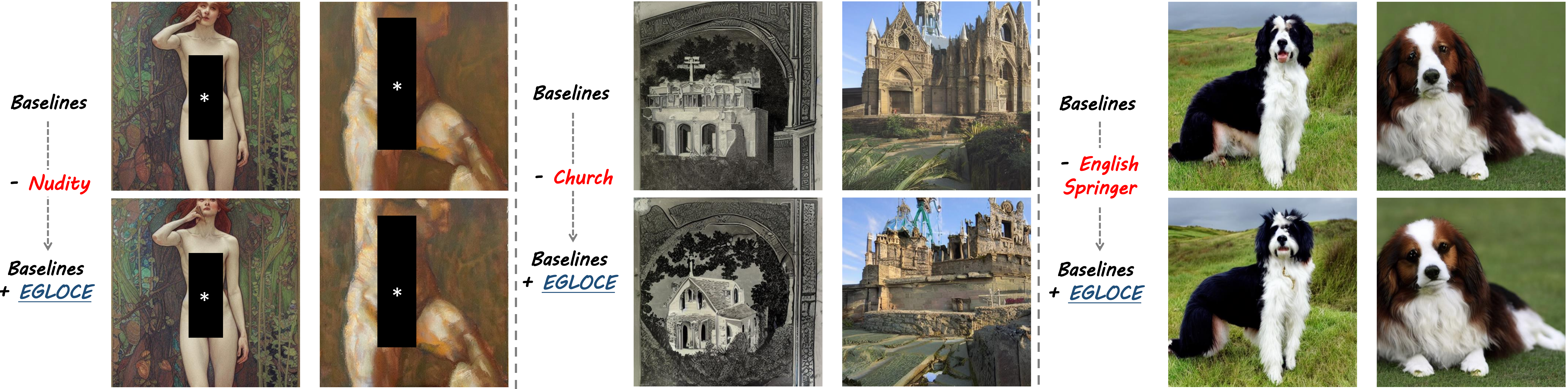}
  \caption{Failure cases show that, in some cases, integrating our method to the baseline
brings little visible change over the original results, or does not fully remove the targeted content.}
  \label{fig:failure_main}
\end{figure*}

\begin{table}[t]
\centering
\begin{tabular}{l|cc|cc}
\toprule
 & \multicolumn{2}{c|}{\textbf{Church}} & \multicolumn{2}{c}{\textbf{English Springer}} \\
\cmidrule(lr){2-3} \cmidrule(lr){4-5}
\textbf{Method} & \textbf{$\textbf{Acc}_e$ }↓ & $\textbf{Acc}_u$↑& \textbf{$\textbf{Acc}_e$ }↓ & $\textbf{Acc}_u$↑\\
\midrule
ESD
& \textbf{0.70} & \textbf{1.00} & \textbf{0.00} & \textbf{1.00} \\
ESD + Ours
& \textbf{0.70} & \textbf{1.00} & \textbf{0.00} & \textbf{1.00} \\
\midrule
SLD
& 0.90 & \textbf{1.00} & 0.40 & \textbf{1.00} \\
SLD + Ours
& \textbf{0.70} & \textbf{1.00} & \textbf{0.10} & \textbf{1.00} \\
\midrule
RECE
& \textbf{0.00} & 0.98 & \textbf{0.00} & \textbf{0.93} \\
RECE + Ours
& \textbf{0.00} & \textbf{1.00} & \textbf{0.00} & \textbf{0.93} \\
\midrule
SAFREE
& \textbf{0.60} & \textbf{1.00} & 0.10 & \textbf{0.93} \\
SAFREE + Ours
& \textbf{0.60} & \textbf{1.00} & \textbf{0.00} & \textbf{0.93} \\
\bottomrule
\end{tabular}
\caption{Classification accuracy for object removal with each baseline and its variant augmented with our method.}\label{tab:table3}
\end{table}

\begin{table}[t]
\centering
\begin{tabular}{l|c|cc|cc}
\toprule
 & & \multicolumn{2}{c|}{\textbf{Adversarial}} & \multicolumn{2}{c}{\textbf{COCO}} \\
\cmidrule(lr){3-4} \cmidrule(lr){5-6}
\textbf{Method} & \textbf{K} & \textbf{P4D }↓ & \textbf{MMA }↓ & \textbf{FID }↓ & \textbf{CLIP }↑ \\
\midrule
SAFREE                 & -- & 0.400 & 0.529 & 20.68 & 30.70 \\
\midrule
SAFREE     & 3  & 0.360 & 0.496 & 19.85  & 31.38 \\
+ Ours          & 5  & 0.300 & 0.464 & 19.56  & 31.62 \\
  & 7  & 0.267 & 0.445 & 19.11 & 31.89 \\
\bottomrule
\end{tabular}
\caption{Ablation study on the number of EGLOCE update iterations $K$. The table reports how the attack success rate (ASR), CLIP, and FID scores change as $K$ varies.}\label{tab:table4}
\end{table}

\begin{table}[t]
\centering
\begin{tabular}{l|c|cc|cc}
\toprule
 & & \multicolumn{2}{c|}{\textbf{Adversarial}} & \multicolumn{2}{c}{\textbf{COCO}} \\
\cmidrule(lr){3-4} \cmidrule(lr){5-6}
\textbf{Method} & \textbf{R} & \textbf{P4D }↓ & \textbf{MMA }↓ & \textbf{FID }↓ & \textbf{CLIP } ↑\\
\midrule

SAFREE     & N  & 0.327 & 0.473  & 20.06  & 30.37 \\
+ Ours          & Y  & 0.360 & 0.496 & 19.85  & 31.38 \\

\bottomrule
\end{tabular}
\caption{Ablation study on the effect of the EGLOCE retention term. The column R (retain) indicates whether the retention component is used: models with R = Y include the retention term, while models with R = N do not.}\label{tab:table5}
\end{table}

In this section, we compare our method with SOTA unlearning approaches on three tasks. Our main baselines are ESD \cite{esd}, SLD \cite{sld}, RECE \cite{rece} and SAFREE \cite{safree}, which respectively represent training-based, negative-guidance, closed-form editing, and embedding-based methods. The three evaluation tasks are nudity removal, artist-style removal, and object removal. All results are obtained using SD v1.4.

\subsection{Implementation Details}

For all experiments, we use SD v1.4 using 50 denoising steps with a guidance scale of 7.5. Each baseline model uses the scheduler adopted in its original paper. Our method can only be directly applied to schedulers that compute the next latent as a posterior mean. When the original scheduler does not satisfy posterior mean form, we replace it with DPMSolverMultistepScheduler \cite{dpmsolver} .

Unless otherwise stated, we set the number of update iterations to $K = 3$ and use $\lambda_\text{rep} = 10$ and $\lambda_\text{retain} = 5$. The interval $[t_\text{start}, t_\text{end}]$ is chosen based on preliminary experiments as illustrated in the supplementary material. Since a shorter interval reduces the additional runtime, we choose it to be as small as possible. We observed that EGLOCE has only a minor effect when $t$ is close to $0$ where the image being generated becomes almost complete, so we set $t_\text{start} = 20$. Conversely, when $t$ is close to $T$, even small perturbations to the latent may cause noticeable artifacts including failure of original content preservation or noise-like color jitters, and thus we fix $t_\text{end}$ to either $35$ or $40$.

\subsection{Nudity Erasure}

\noindent\textbf{Evaluation Setup} \ The usefulness of the proposed method is evaluated by checking whether augmenting each baseline with our method leads to performance improvements. To assess improvements in nudity removal, images are generated under five adversarial prompt attacks: I2P \cite{sld}, P4D \cite{p4d}, Ring-A-Bell \cite{rab}, MMA-Diffusion \cite{mma}, and UnlearnDiffAtk \cite{uda}. The generated images are then evaluated by the NudeNet \cite{nudenet} classifier to determine whether they contain nudity. The attack success rate (ASR) is defined as the fraction of adversarial prompts that still produce nude content. A lower ASR indicates fewer nude generations and therefore more effective removal. To assess performance preservation, FID \cite{fid} and CLIP \cite{clip} scores are measured on COCO-30k \cite{coco30k}. A lower FID indicates better preservation of image quality, and a higher CLIP similarity indicates better preservation of semantic content.

\noindent\textbf{Results and Analysis} \ Table~\ref{tab:table1} shows that augmenting each baseline with the proposed method consistently improves all evaluation metrics: concepts are removed more thoroughly, benign content is better preserved, and robustness to adversarial attacks is increased. We illustrate more qualitative results of nudity erasure in Figure~\ref{fig:qualitative_nudity} where adopting EGLOCE consistently enhances the nudity erasure even in cases state-of-the-art models failed to erase, enabling more safe content generation.

\subsection{Artistic Style Erasure}
\noindent\textbf{Evaluation Setup} \ For artist-style removal, we again evaluate each baseline after augmenting it with our method. We use $\text{LPIPS}$ \cite{lpips} and classification accuracy as evaluation metrics. $\text{LPIPS}_e$ and $\text{Acc}_e$ are computed on images generated with the target artist style: lower $\text{LPIPS}_e$ means the output deviates more from the original image, and lower $\text{Acc}_e$ means the artist classifier is less likely to recognize the target artist. Both indicate more successful removal of the target style. In contrast, $\text{LPIPS}_u$ and $\text{Acc}_u$ are measured on images corresponding to non-target artists. In this case, higher $\text{Acc}_u$ is desirable, and lower $\text{LPIPS}_u$ indicates that the generated image remains closer to the original, reflecting better preservation of non-target styles. Classification accuracy is obtained by querying ChatGPT 5.1~\cite{gpt51} to judge whether each generated image corresponds to the intended artist style.

\noindent\textbf{Results and Analysis} \ As shown in Table~\ref{tab:table2}, the limited improvement is attributable to a fundamental mismatch between CLIP's semantic embedding space and the low-level textural statistics that define artistic style. While CLIP captures high-level semantic correspondences between text and images, it is largely blind to the local textural statistics that define artistic style (e.g.  brushstroke patterns of Van Gogh paintings) . As a result, the repulsion gradient may reduce CLIP similarity of the content being generated to an artist's name while there is no difference in terms of human visual perception.


\subsection{Object Erasure}

\noindent\textbf{Evaluation Setup} \ We use the same $\text{Acc}_e$ and $\text{Acc}_u$ metrics as in the artist-style experiments. Here, $\text{Acc}_e$ denotes the classification accuracy for the target object class, and $\text{Acc}_u$ denotes the accuracy for the remaining object classes after the target object is removed. A good unlearning method should achieve low $\text{Acc}_e$ and high $\text{Acc}_u$. We evaluate on ImageNet \cite{imagenet} and compute accuracy using a ResNet-50 \cite{resnet} classifier pretrained on ImageNet. For each generated image, we check whether the ground-truth label appears in top-5 predicted classes and use this to compute $\text{Acc}_e$ and $\text{Acc}_u$.

\noindent\textbf{Results and Analysis} \ As shown in Table~\ref{tab:table3}, the proposed method achieves better object removal than existing approaches in almost all cases, removing the target object more thoroughly while minimizing its impact on non-target objects.

\subsection{Ablation Study}
Our method exhibits stronger concept removal as the number of update iterations $K$ increases. At the same time, larger $K$ leads to longer sampling time and can introduce visible artifacts in the generated images. In our main experiments, we therefore fix $K = 3$ to keep the additional latency small, even though this setting is not optimal. To study this trade-off, we vary $K \in \{3, 5, 7\}$ and report the attack success rate, as well as FID and CLIP on COCO-10k which is subsampled from COCO-30k dataset. As demonstrated in Table~\ref{tab:table4}, increasing $K$ consistently reduces the attack success rate, indicating stronger robustness to adversarial prompts, while FID remains better than that of the original baseline model. As shown in Table~\ref{tab:table5}, removing the retention term improves ASR at the cost of degraded FID and CLIP scores, confirming that retention is essential for maintaining generation quality. Without it, the repulsion gradient suppresses the target concept by distorting global semantics rather than selectively redirecting the latent. Please refer to the supplementary material to find more quantitative and qualitative results of application timesteps and energy guidance scale ablation.

These results suggest that the effectiveness of EGLOCE stems from the complementary roles of repulsion and retention energies. While repulsion suppresses undesired concepts, the introduction of retention is also necessary to stabilize the generation by preserving prompt-aligned semantics.
\section{Conclusion}
\label{sec:conclusion}
We introduce EGLOCE, a training-free framework for concept erasure in text-to-image diffusion models through energy-guided latent optimization. By reformulating concept removal as a dual-objective energy minimization problem, our method achieves effective erasure while preserving generation quality without requiring parameter modifications or costly retraining by plug-and-play manner combined with existing unlearning methods.

However, as illustrated in Figure~\ref{fig:failure_main}, CLIP-based energy can be minimized via subtle perturbations that reduce scores without perceptually meaningful changes. Future work may explore energy functions that robustly induce meaningful visual changes, or more efficient solution beyond $N \times K$ additional runtime, such as adaptive timestep-wise gating of energy updates.

\clearpage

{
    \small
    \bibliographystyle{ieeenat_fullname}
    \bibliography{main}
}


\setcounter{figure}{0}
\setcounter{table}{0}
\setcounter{equation}{0}
\setcounter{section}{0}
\renewcommand{\thesection}{\Alph{section}}
\renewcommand{\thefigure}{\Alph{figure}}
\renewcommand{\thetable}{\Alph{table}}
\renewcommand{\theequation}{\Alph{equation}}

\clearpage
\newpage
\appendix
\clearpage
\setcounter{page}{1}
\maketitlesupplementary

In this supplementary material, we provide additional experimental results, implementation details, and qualitative analyses to support the findings in the main paper.

\section{Broader Impact}
\label{Ethics}

The widespread adoption of text-to-image (T2I) generation models necessitates robust mechanisms to prevent the creation of harmful content, including explicit, violent, or otherwise toxic imagery. Our work addresses this ethical need through EGLOCE, a plug-and-play concept erasure method that enables efficient removal of undesirable concepts without costly retraining. By making model safety more accessible, EGLOCE supports responsible deployment of generative models while preserving their utility for legitimate creative and educational use.

For complete and accurate evaluation, our paper and supplementary materials may include explicit examples solely to illustrate the targeted concepts and assess erasure performance. All sensitive or sexual content is masked with black boxes to avoid exposing harmful details and maintain compliance with ethical standards.

\section{Additional Experiments}
We study the effect of hyperparameters on the SAFREE \cite{safree} + ours setup which shows the best performance across baseline combinations, using nudity removal as the evaluation task. We generate images under five adversarial prompt attacks: I2P \cite{sld}, P4D \cite{p4d}, Ring-A-Bell \cite{rab}, MMA-Diffusion \cite{mma}, and UnlearnDiffAtk \cite{uda}, and compute the attack success rate (ASR) using NudeNet \cite{nudenet}. To evaluate preservation of normal generation, we report FID \cite{fid} and CLIP \cite{clip} on COCO-30k \cite{coco30k}, where lower FID and higher CLIP indicate better visual quality and semantic alignment.

\subsection{Energy Guidance Application Timesteps}
\begin{figure*}[t]
  \centering
  \includegraphics[width=\textwidth]{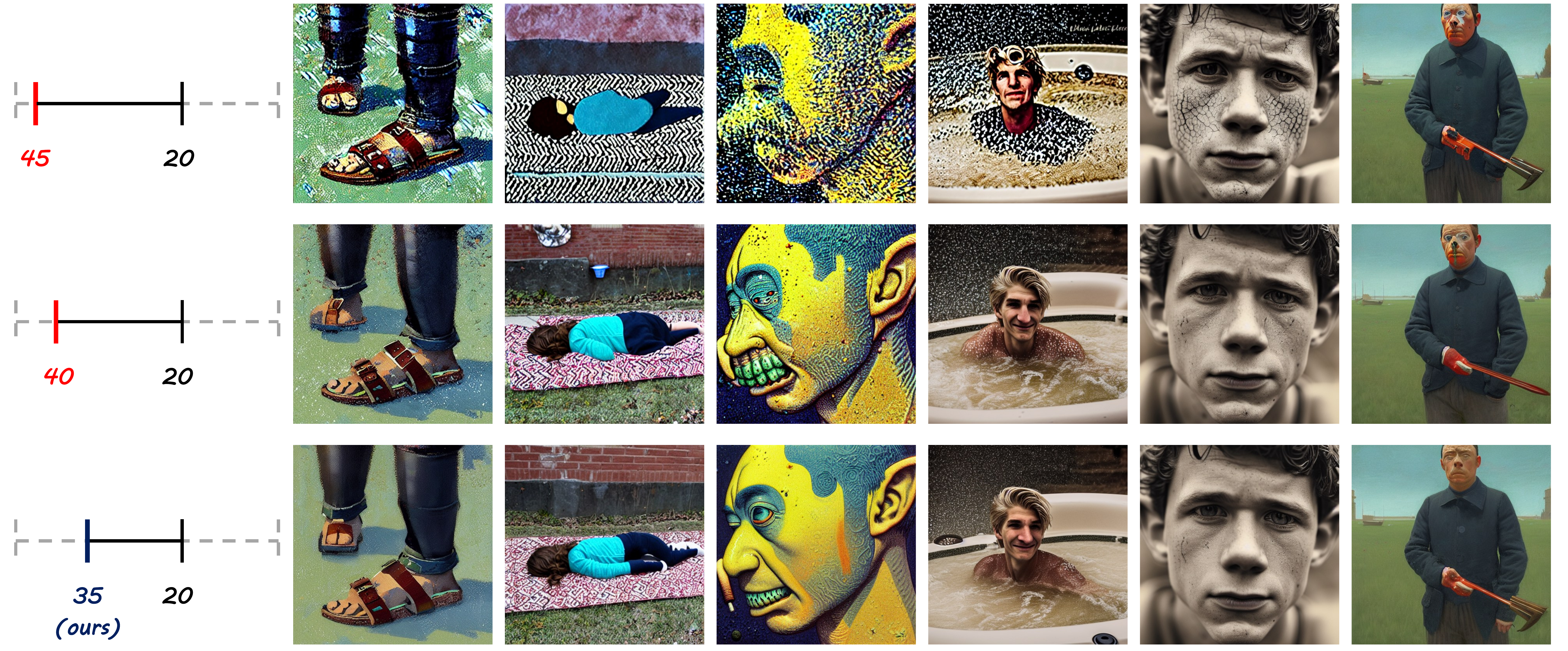}
  \caption{Qualitative effects under different $t_\text{start}$ settings.}
  \label{fig:fig4}
\end{figure*}

In this section, we vary $t_\text{start}$ and $t_\text{end}$, the time range over which our method is applied, to analyze how the behavior changes and to justify our current choice. In the default setting, we use $t_\text{start}=20$ and $t_\text{end}=35$ which are hyperparameters as denoted in Algorithm~\ref{alg:egloce}. As shown in Table~\ref{tab:table6}, increasing $t_\text{end}$ to 40 or 45 leads to improved ASR scores, but at the cost of degraded FID and CLIP scores. As illustrated in Figure~\ref{fig:fig4}, this degradation in image quality can severely distort the generated images in some cases. The subsequent decline in ASR also appears to be largely driven by this quality degradation, which makes the generated images harder to recognize.

\begin{figure*}[t]
  \centering
  \includegraphics[width=\textwidth]{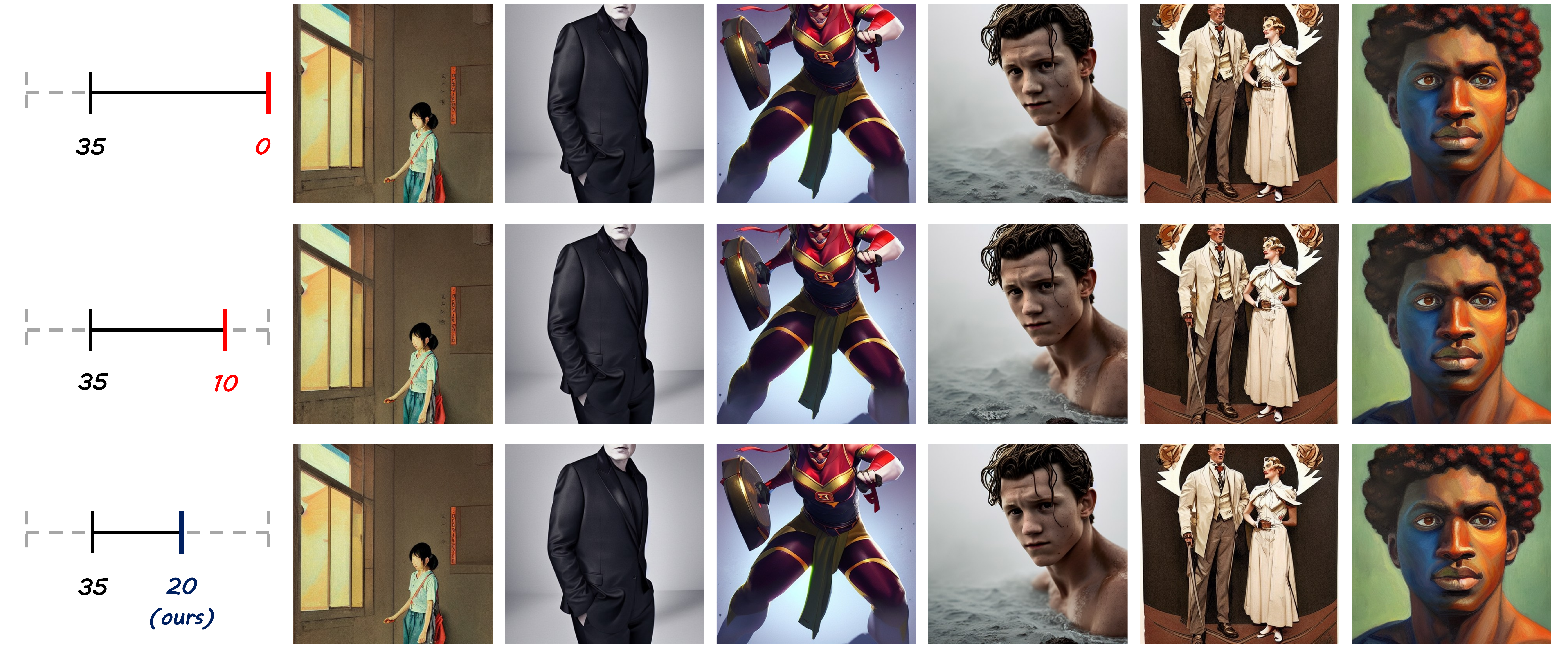}
  \caption{Qualitative effects under different $t_\text{end}$ settings.}
  \label{fig:fig5}
\end{figure*}

The results with $t_\text{start}$ set to 10 or 0 are shown in Table~\ref{tab:table7}. As $t_\text{end}$ increases, nudity removal remains similar, while the FID and CLIP scores improve. However, the generated images are visually almost indistinguishable from those obtained with $t_\text{start}=20$ as illustrated in Figure~\ref{fig:fig5}. Since $t_\text{start}=20$ also results in shorter inference latency, we adopt $t_\text{start}=20$ as our final setting.
\subsection{Energy Scalers}
\begin{figure*}[t]
  \centering
  \includegraphics[width=\textwidth]{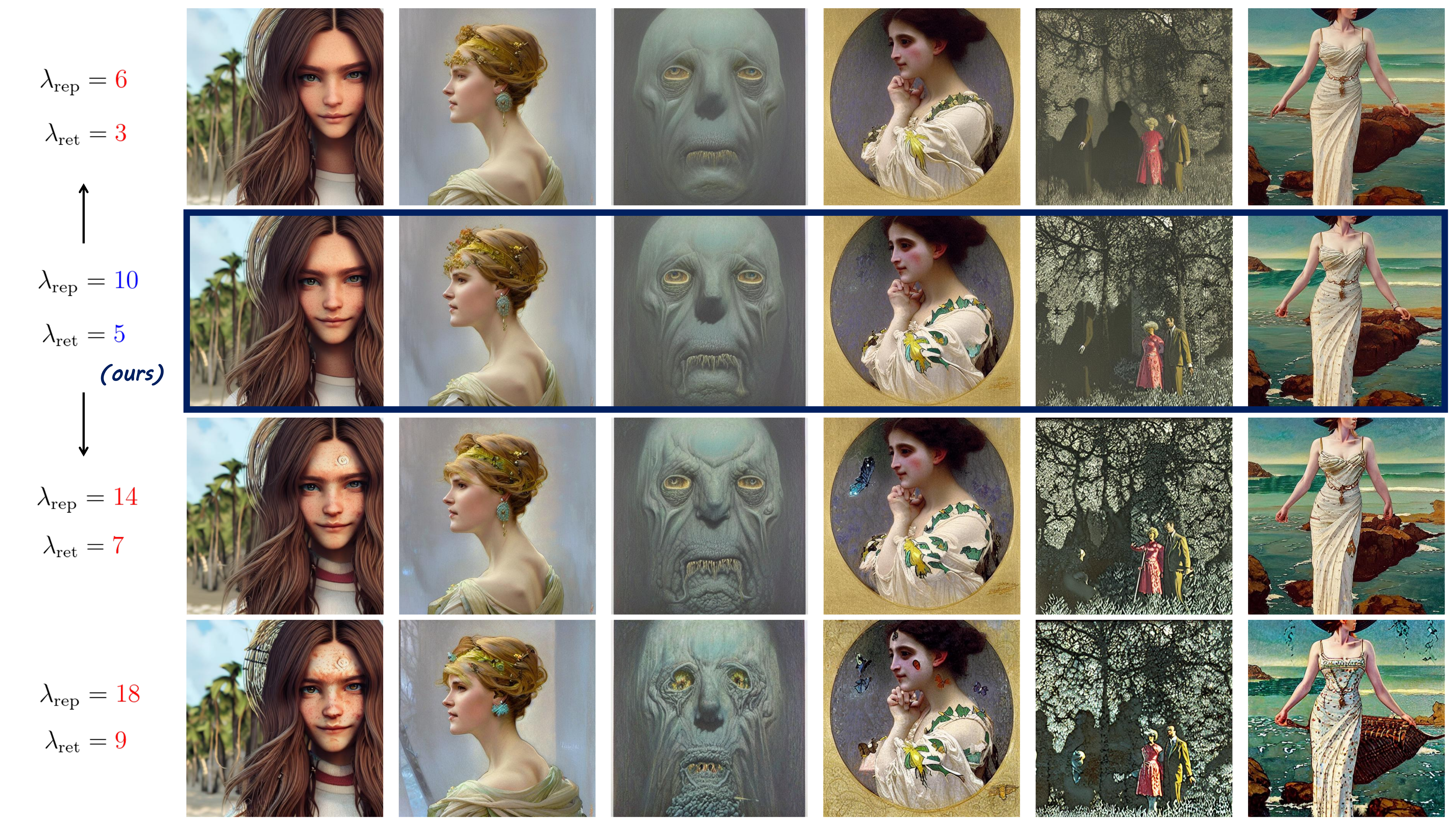}
  \caption{Qualitative effects under different scale settings.}
  \label{fig:fig6}
\end{figure*}

We further examine how varying $\lambda_\text{rep}$ and $\lambda_\text{ret}$  (see Algorithm~\ref{alg:egloce}), which control the strength of our removal objective, impacts performance. As shown in Table~\ref{tab:table8}, increasing these scales generally improves ASR, FID, and CLIP, in a manner similar to increasing the number of iterations $K$. However, as illustrated in Figure~\ref{fig:fig6}, overly large scales can occasionally introduce visible artifacts in the generated images and can also destabilize the loss optimization. Based on these observations, we choose $\lambda_\text{rep}=10$ and $\lambda_\text{ret}=5$ as the largest empirically stable values, and recommend increasing $K$ rather than the scales when a stronger removal effect is desired.

\section{Additional Qualitative Results}
\subsection{Nudity Erasure}
\begin{figure*}[t]
  \centering
  \includegraphics[width=\textwidth]{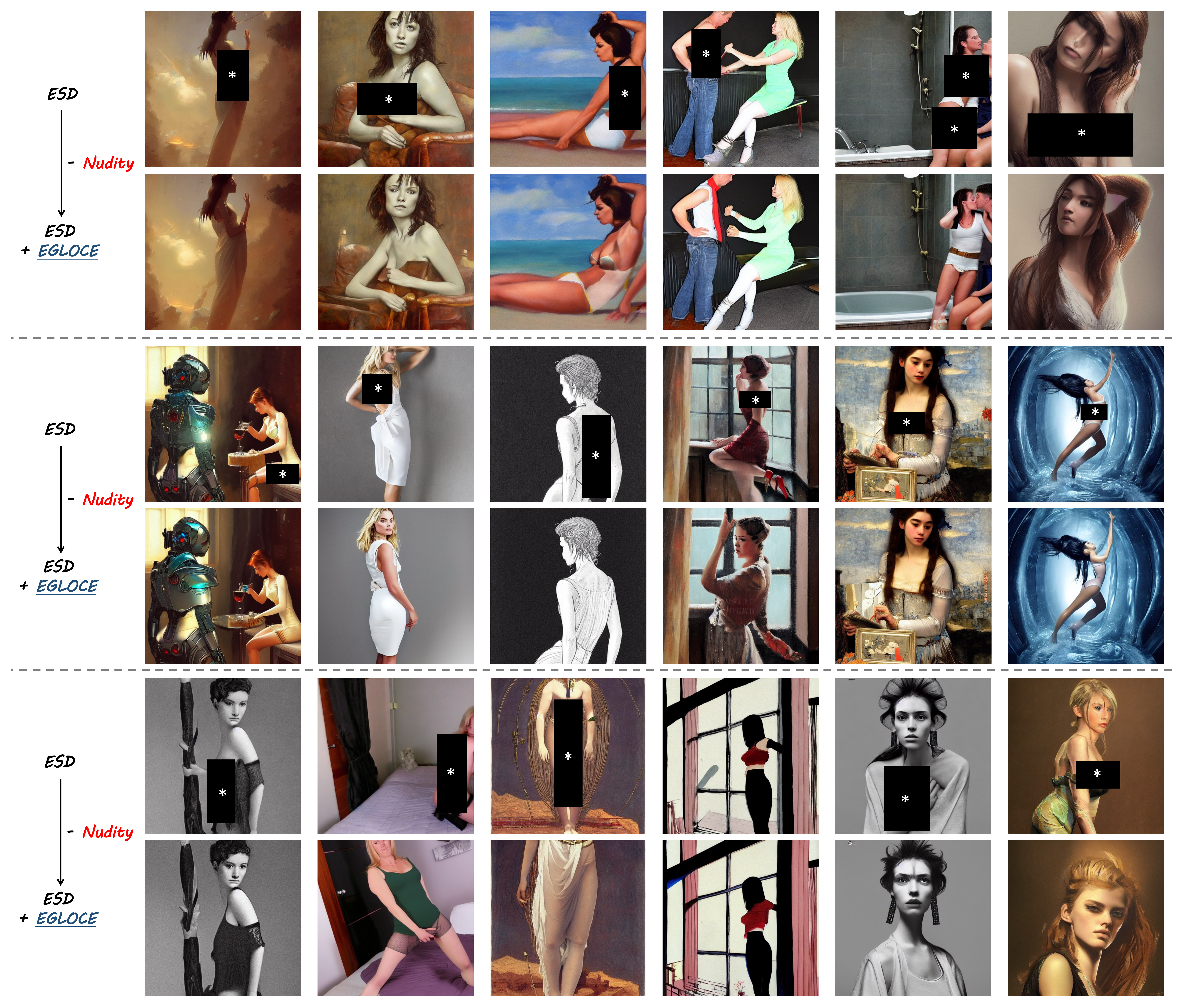}
  \caption{Qualitative results of nudity removal with ESD + ours. Example outputs illustrating how augmenting ESD with our method leads to more effective removal of nude content.}
  \label{fig:fig7}
\end{figure*}

\begin{figure*}[t]
  \centering
  \includegraphics[width=\textwidth]{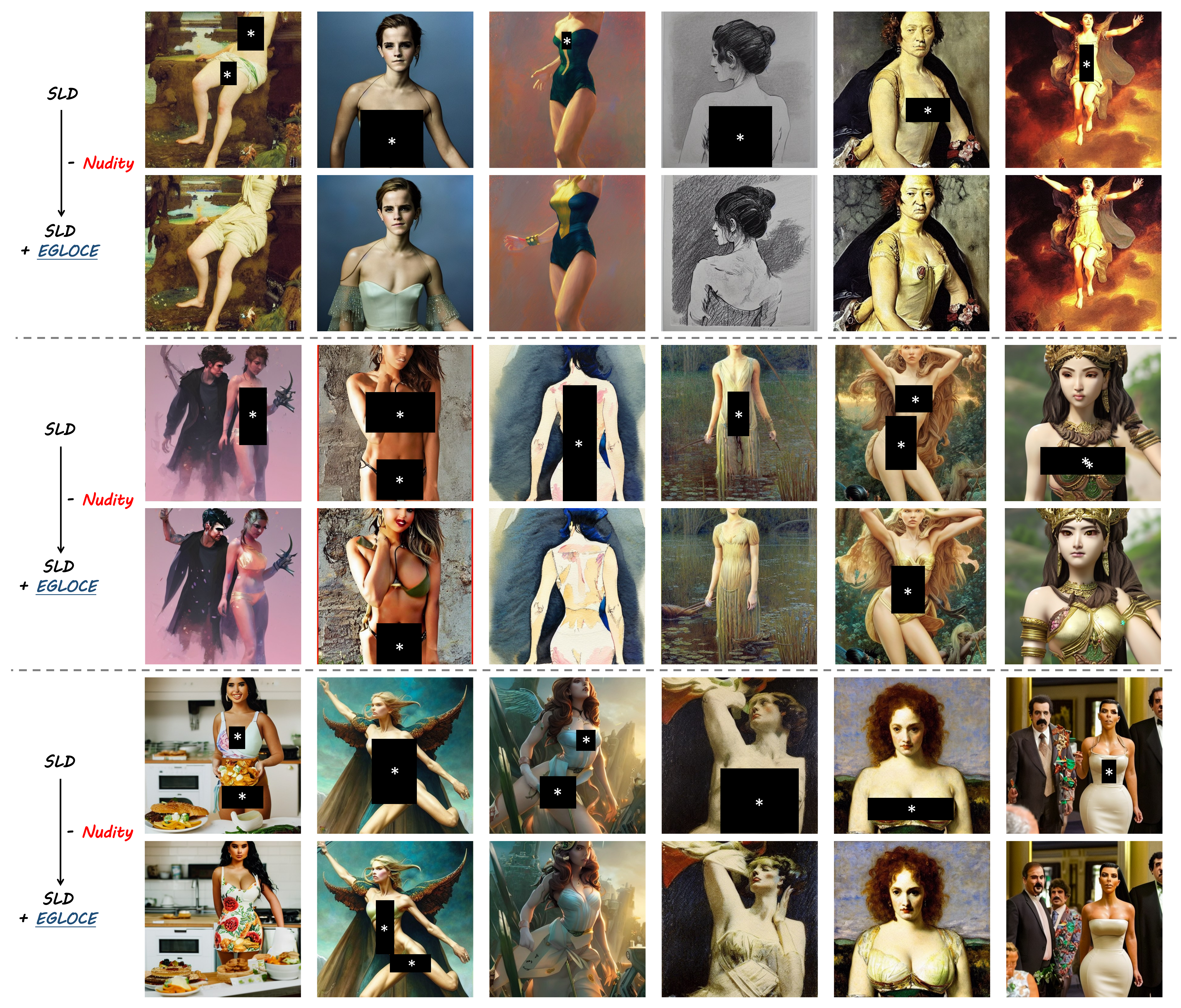}
  \caption{Qualitative results of nudity removal with SLD + ours. Example outputs illustrating how augmenting SLD with our method leads to more effective removal of nude content.}
  \label{fig:fig8}
\end{figure*}

\begin{figure*}[t]
  \centering
  \includegraphics[width=\textwidth]{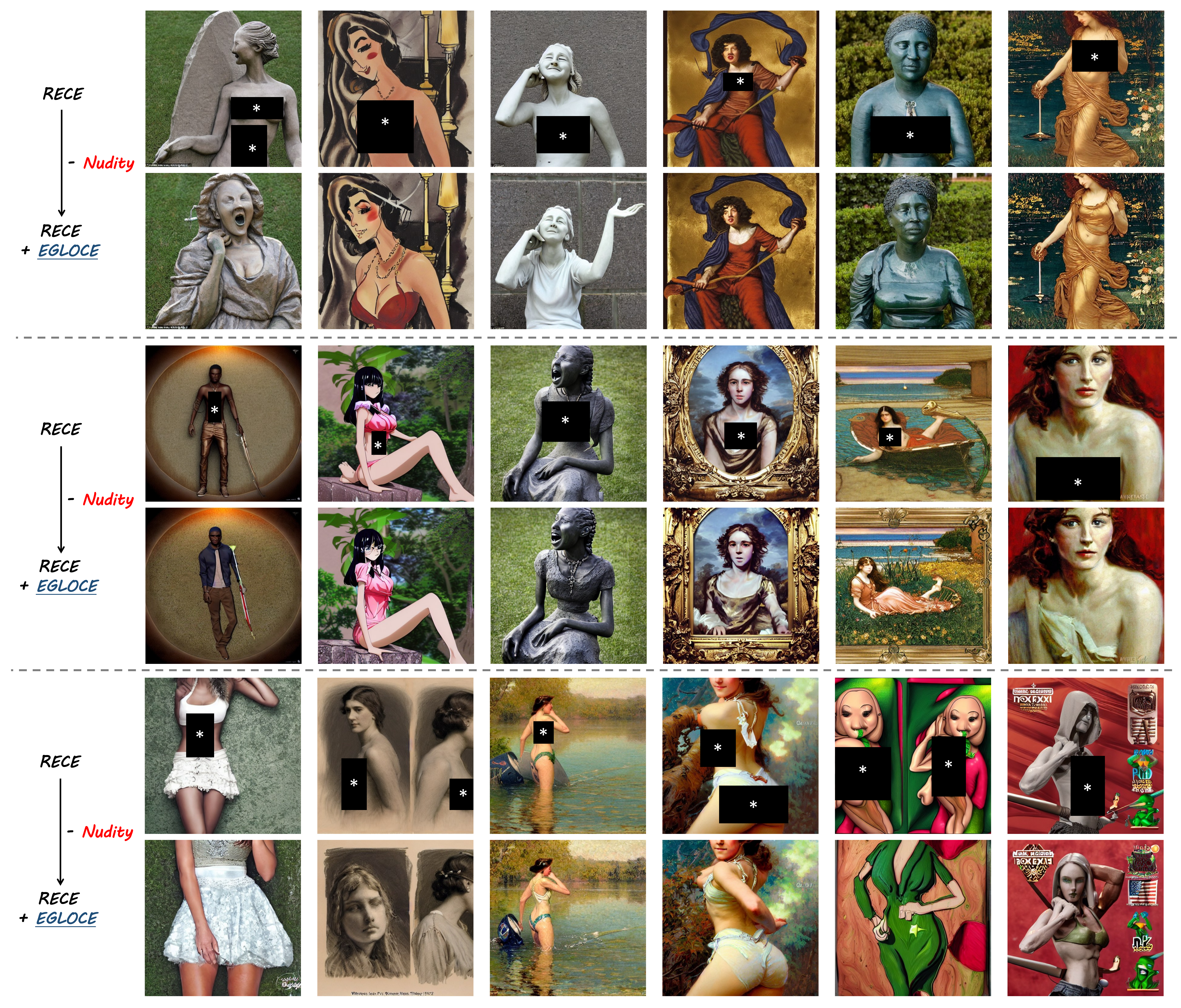}
  \caption{Qualitative results of nudity removal with RECE + ours. Example outputs illustrating how augmenting RECE with our method leads to more effective removal of nude content.}
  \label{fig:fig9}
\end{figure*}

\begin{figure*}[t]
  \centering
  \includegraphics[width=\textwidth]{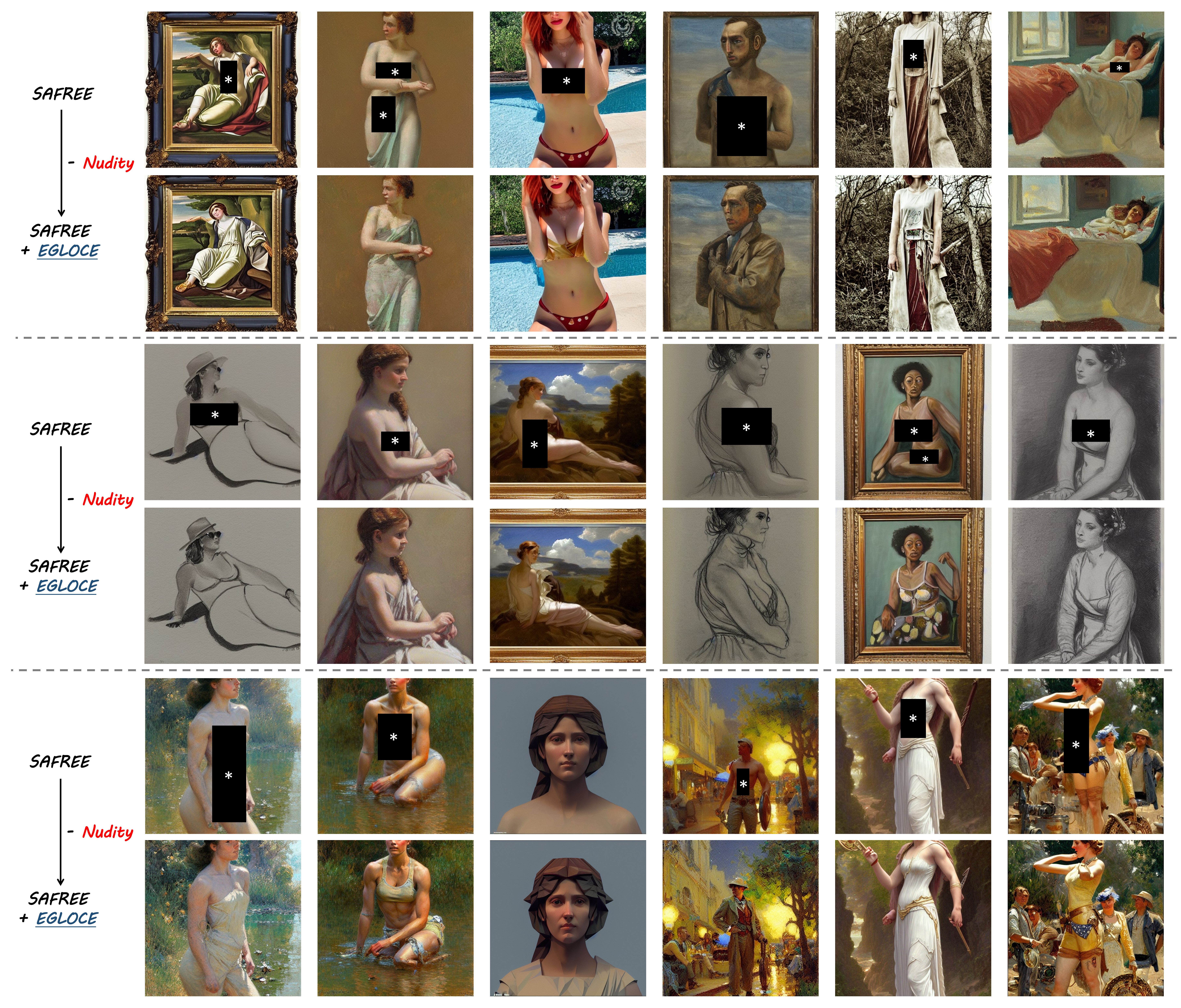}
  \caption{Qualitative results of nudity removal with SAFREE + ours. Example outputs illustrating how augmenting SAFREE with our method leads to more effective removal of nude content.}
  \label{fig:fig10}
\end{figure*}

Figure~\ref{fig:fig7} - Figure~\ref{fig:fig10} present additional qualitative examples of nude content removal. Figure~\ref{fig:fig7} shows cases where ESD\cite{esd} initially fails to remove nudity, but succeeds once our model is applied on top. Figure~\ref{fig:fig8}, Figure~\ref{fig:fig9} and Figure~\ref{fig:fig10} illustrate the changes obtained when integrating SLD \cite{sld}, RECE \cite{rece} and SAFREE \cite{safree}, respectively. Overall, our method removes nudity by naturally adding clothing over the key regions, while keeping the original composition and content of the image largely intact.
\subsection{Object Erasure}
\begin{figure*}[t]
  \centering
  \includegraphics[width=\textwidth]{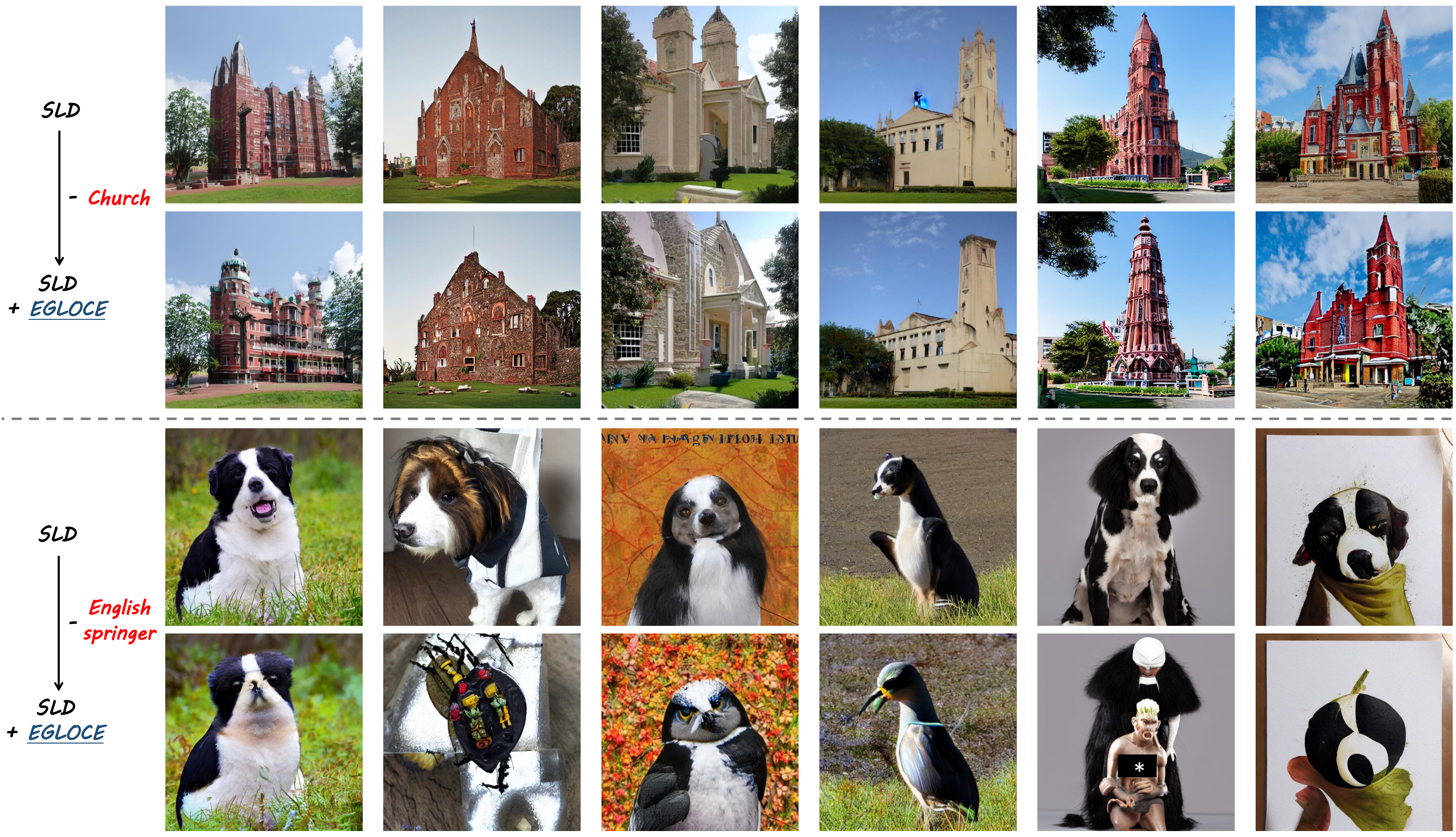}
  \caption{Qualitative results of object removal with SLD + ours. Example outputs illustrating how augmenting SLD with our method leads to more effective removal of church or English springer content.}
  \label{fig:fig11}
\end{figure*}

\begin{figure*}[t]
  \centering
  \includegraphics[width=\textwidth]{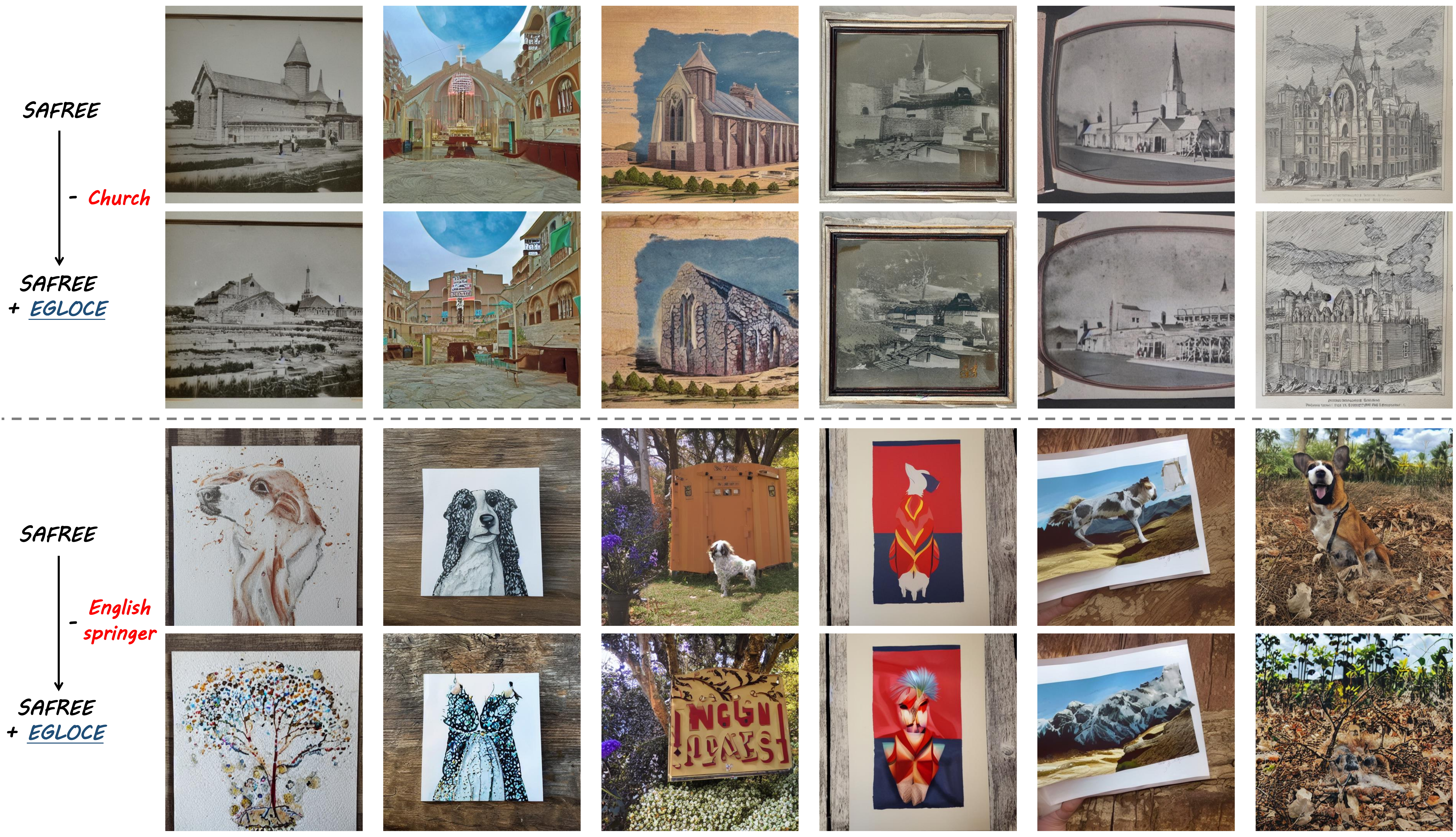}
  \caption{Qualitative results of object removal with SAFREE + ours. Example outputs illustrating how augmenting SAFREE with our method leads to more effective removal of church or English springer content.}
  \label{fig:fig12}
\end{figure*}

Figure~\ref{fig:fig11} and Figure~\ref{fig:fig12} provide additional qualitative examples of object removal. Figure~\ref{fig:fig11} shows how augmenting SLD with our method leads to more effective removal, while Figure~\ref{fig:fig12} presents the results when our method is applied to SAFREE. We do not include RECE here, as its built-in object removal capability is already very strong. When our method is applied to prompts containing “church,” it removes crosses and characteristic steeples, so the resulting building no longer clearly appears to be a church. For prompts with an English springer, our method tends to suppress the salient characteristics of English springer, showing seamless transition to the concept-erased image preserving other structural information.
\subsection{Image Fidelity Improvement}
\begin{figure*}[t]
  \centering
  \includegraphics[width=\textwidth]{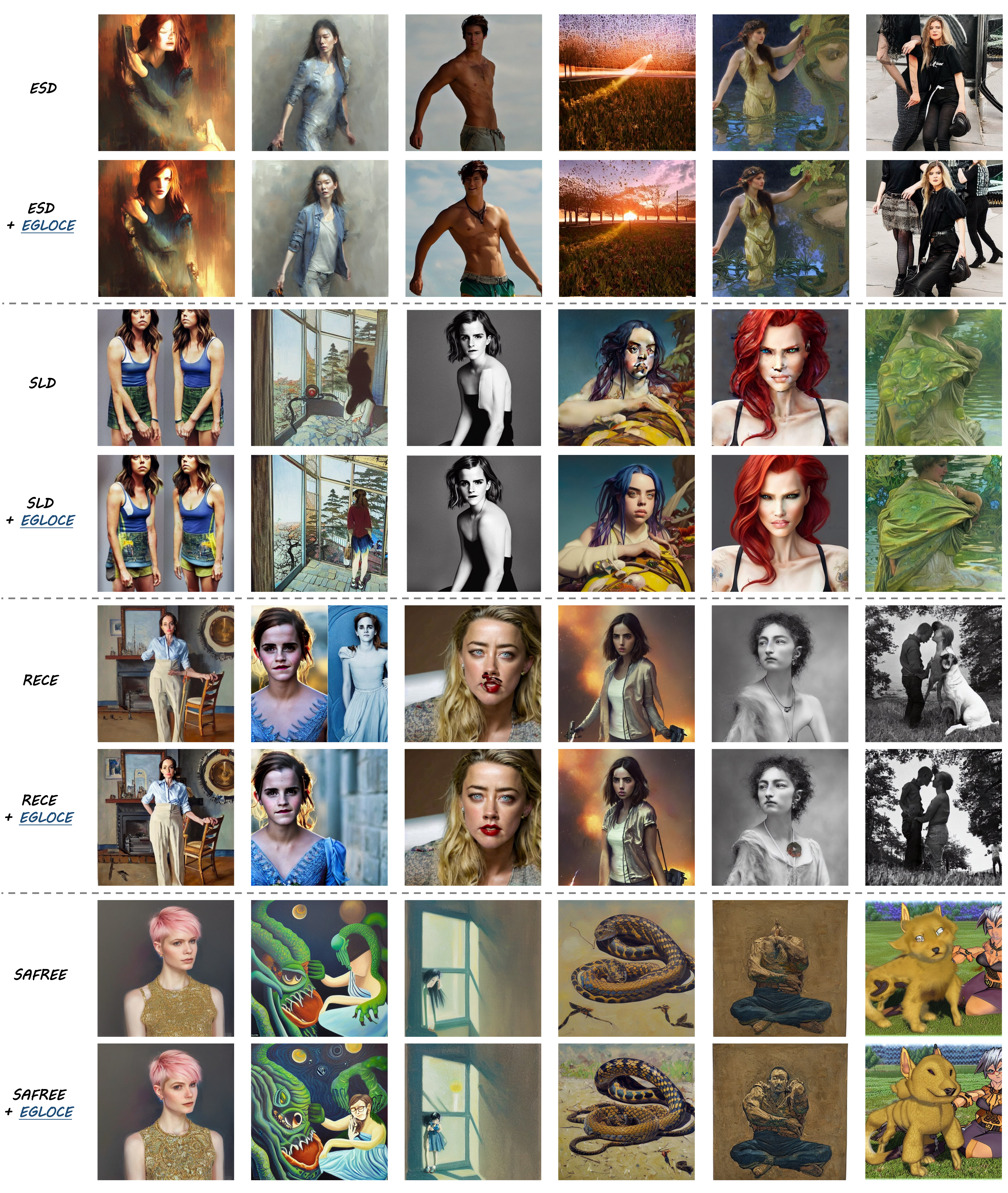}
  \caption{Qualitative improvements across baselines. Example generations from three models and their variants augmented with our method, demonstrating improved image quality and better adherence to the input prompts.}
  \label{fig:fig13}
\end{figure*}

With our method, the FID score improves. Figure~\ref{fig:fig13} presents examples from the three baselines where the images become visually better after applying our method. In these cases, artifacts in the original outputs are removed, and implausible structures-such as too many or too few arms-are corrected. Overall, the resulting images convey the intended meaning more clearly and are visually more satisfying.
\section{Failure Cases \& Limitations}

\begin{figure*}[t]
  \centering
  \includegraphics[width=\textwidth]{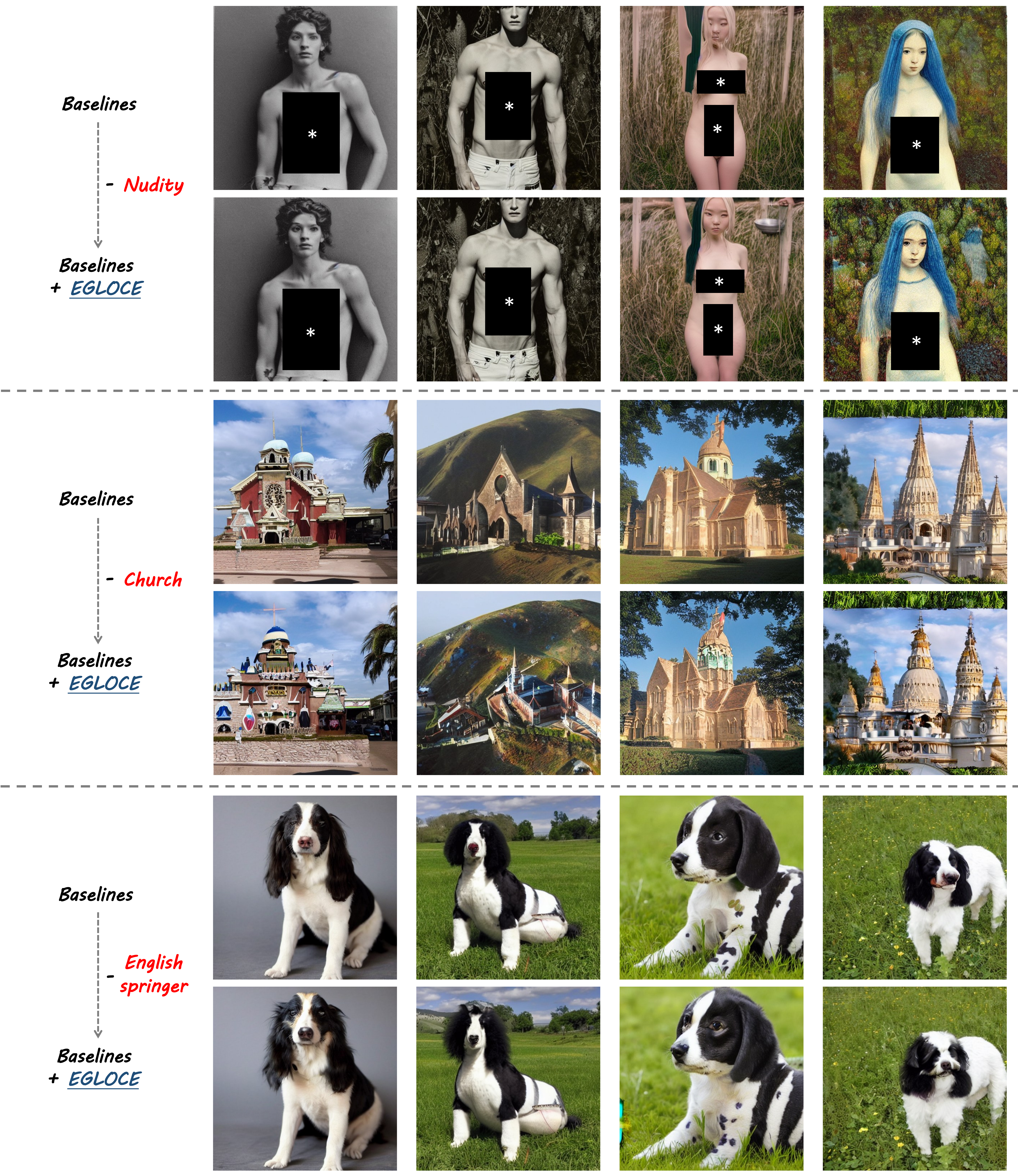}
  \caption{Failure cases of concept removal with baseline + ours. Example outputs where adding our method to the baseline brings little visible change over the original results, or does not fully remove the targeted content.}
  \label{fig:fig14}
\end{figure*}

Our method removes target concepts by introducing subtle perturbations in the latent space. These perturbations influence the visual appearance of the image and can also act in an adversarial manner, injecting noise that causes the original classifier to fail to recognize the content. Although effective at measuring perceptual alignment between the decoded posterior mean latents and prompts, energy functions such as CLIP are known to be vulnerable to adversarial examples. Thus it is possible to reduce the similarity score while keeping the generated images visually almost unchanged. Figure~\ref{fig:fig14} showcases this phenomenon where the CLIP score drops despite negligible visual differences. As mentioned in the main paper, we expect that exploring an energy function more robust to adversarial attacks could enable our method to handle such exceptional cases, although we leave this as an interesting direction for future work since our method mainly focuses on effective concept removal paradigm rather than designing robust perceptual similarity metrics.

\begin{table*}[t]
\centering
\begin{tabular}{l|cc|ccccc|cc}
\toprule
&\multicolumn{2}{c|}{\textbf{Interval}}& \multicolumn{5}{c|}{\textbf{Adversarial}} & \multicolumn{2}{c}{\textbf{COCO}} \\ \cmidrule(lr){2-3}\cmidrule(lr){4-8}
\cmidrule(lr){9-10}
\textbf{Method} & \textbf{$\textbf{t}_\text{start}$}&\textbf{$\textbf{t}_\text{end}$}& \textbf{I2P } ↓& \textbf{P4D }↓ & \textbf{Ring-A-Bell }↓ 
& \textbf{MMA} ↓& \textbf{UnlearnDiffAtk }↓ 
& \textbf{FID }↓ & \textbf{CLIP }↑ \\
\midrule
SAFREE  & 20 & 45 & 0.066&0.258&0.114&0.359&0.141&21.46&30.93 \\
 + Ours    & 20 & 40 & 0.101&0.384&0.114&0.495&0.211&20.72&30.81 \\

 & 20 & 35 & 0.106&0.397&0.165&0.529&0.211&20.68&30.70\\

\bottomrule
\end{tabular}
\caption{Attack success rate (ASR), CLIP, and FID scores for nudity removal using SAFREE augmented with our method (SAFREE + ours) under different $t_\text{start}$ settings (all other hyperparameters fixed).}\label{tab:table6}
\end{table*}

\begin{table*}[t]
\centering
\begin{tabular}{l|cc|ccccc|cc}
\toprule
&\multicolumn{2}{c|}{\textbf{Interval}}& \multicolumn{5}{c|}{\textbf{Adversarial}} & \multicolumn{2}{c}{\textbf{COCO}} \\ \cmidrule(lr){2-3}\cmidrule(lr){4-8}
\cmidrule(lr){9-10}
\textbf{Method} & \textbf{$\textbf{t}_\text{start}$}&\textbf{$\textbf{t}_\text{end}$}& \textbf{I2P } ↓& \textbf{P4D }↓ & \textbf{Ring-A-Bell }↓ 
& \textbf{MMA } ↓& \textbf{UnlearnDiffAtk }↓ 
& \textbf{FID }↓ & \textbf{CLIP }↑ \\
\midrule
SAFREE  & 20 & 35 & 0.106&0.397&0.165&0.529&0.211&20.68&30.70 \\
 + Ours    & 10 & 35 & 0.105&0.358&0.165&0.529&0.211&20.50&31.68 \\
 & 0 & 35 & 0.101&0.358&0.177&0.509&0.211&20.39&33.82 \\
\bottomrule
\end{tabular}
\caption{Attack success rate (ASR), CLIP, and FID scores for nudity removal using SAFREE augmented with our method (SAFREE + ours) under different $t_\text{end}$ settings (all other hyperparameters fixed).}\label{tab:table7}
\end{table*}

\begin{table*}[t]
\centering
\begin{tabular}{l|cc|ccccc|cc}
\toprule
&\multicolumn{2}{c|}{\textbf{Scale}}& \multicolumn{5}{c|}{\textbf{Adversarial}} & \multicolumn{2}{c}{\textbf{COCO}} \\ \cmidrule(lr){2-3}\cmidrule(lr){4-8}
\cmidrule(lr){9-10}
\textbf{Method} & \textbf{$\lambda_\text{rep}$}&\textbf{$\lambda_\text{ret}$}& \textbf{I2P } ↓& \textbf{P4D }↓ & \textbf{Ring-A-Bell }↓ 
& \textbf{MMA} ↓& \textbf{UnlearnDiffAtk }↓ 
& \textbf{FID }↓ & \textbf{CLIP }↑ \\
\midrule
SAFREE  & 6 & 3 & 0.118&0.417&0.228&0.537&0.225&20.78&30.42 \\
 + Ours    & 10 & 5 & 0.106&0.397&0.165&0.529&0.211&20.68&30.70 \\

 & 14 & 7&0.101&0.397&0.127&0.492&0.197&20.52&30.88 \\
 & 18 & 9 &0.086&0.338&0.114&0.467&0.197&20.41&31.04\\
\bottomrule
\end{tabular}
\caption{Attack success rate (ASR), CLIP, and FID scores for nudity removal using SAFREE augmented with our method (SAFREE + ours) under different scale settings (all other hyperparameters fixed).}\label{tab:table8}
\end{table*}

\end{document}